\documentclass[lettersize,journal]{IEEEtran}
\usepackage{amsmath,amsfonts}
\usepackage{algorithm}
\usepackage{algpseudocode}
\usepackage{array}
\usepackage{textcomp}
\usepackage{stfloats}
\usepackage{url}
\usepackage{verbatim}
\usepackage{graphicx}
\usepackage{cite}
\usepackage{hyperref}
\usepackage{xcolor}
\usepackage{harpoon}
\usepackage{mathtools}
\usepackage{multirow}
\usepackage{enumitem}
\usepackage{float}
\usepackage{array}
\usepackage{makecell}

\usepackage[font=small]{caption}
\usepackage[font=scriptsize,labelformat=simple]{subcaption}

\makeatletter
\begin{document}

\title{Goal-oriented Communication for Fast and Robust Robotic Fault Detection and Recovery}

\author{Shutong Chen,
Adnan Aijaz,~\IEEEmembership{Senior Member,~IEEE}, 
and Yansha Deng,~\IEEEmembership{Senior Member,~IEEE}
\thanks{S. Chen and Y. Deng are with the Department of Engineering, King’s College London, London WC2R 2LS, U.K. (e-mail: shutong.chen@kcl.ac.uk; yansha.deng@kcl.ac.uk) (Corresponding author: Yansha Deng).

A. Aijaz is with the Bristol Research and Innovation Laboratory, Toshiba Europe Ltd., Avon, Bristol, BS1 4ND,  U.K. (e-mail: adnan.aijaz@toshiba.eu).
}}


\maketitle

\begin{abstract}
Autonomous robotic systems are widely deployed in smart factories and operate in dynamic, uncertain, and human-involved environments that require low-latency and robust fault detection and recovery (FDR).
However, existing FDR frameworks exhibit various limitations, such as significant delays in communication and computation, and unreliability in robot motion/trajectory generation, mainly because the communication-computation-control (3C) loop is designed without considering the downstream FDR goal.
To address this, we propose a novel Goal-oriented Communication (GoC) framework that jointly designs the 3C loop tailored for fast and robust robotic FDR, with the goal of minimising the FDR time while maximising the robotic task (e.g., workpiece sorting) success rate.
For fault detection, our GoC framework innovatively defines and extracts the 3D scene graph (3D-SG) as the semantic representation via our designed representation extractor, and detects faults by monitoring spatial relationship changes in the 3D-SG. 
For fault recovery, we fine-tune a small language model (SLM) via Low-Rank Adaptation (LoRA) and enhance its reasoning and generalization capabilities via knowledge distillation to generate recovery motions for robots.
We also design a lightweight goal-oriented digital twin reconstruction module to refine the recovery motions generated by the SLM when fine-grained robotic control is required, using only task-relevant object contours for digital twin reconstruction.
Extensive simulations demonstrate that our GoC framework reduces the FDR time by up to 82.6\% and improves the task success rate by up to 76\%, compared to the state-of-the-art frameworks that rely on vision language models for fault detection and large language models for fault recovery.
Demo link: \url{https://sites.google.com/view/gscfdr}.
\end{abstract}

\begin{IEEEkeywords}
Goal-oriented communication, semantic representation, robot replan, scene graph, point cloud sampling, digital twin, small language model, and knowledge distillation.
\end{IEEEkeywords}

\section{Introduction}
Autonomous robotic systems have been widely deployed in various industrial applications, such as parcel sorting, box packaging, and machine loading. 
These multi-stage and long-horizon tasks require robots to operate in dynamic and partially observable environments without supervision.
However, a high level of autonomy also amplifies system fragility, where random external disturbances (e.g., collisions) or execution errors (e.g., grasp failures) can lead to control deviations, ultimately reducing operational efficiency and even posing safety risks.  
For example, in human-robot collaboration, the average human reaction time to unexpected robot behaviours is approximately  500 ms \cite{9981726}, yet the time-to-contact of a robotic collision event is about 200 ms \cite{collision}, which means the robot may have already made unintended contact  with its surroundings before a human operator can respond effectively.
This mismatch motivates the need for a fast and robust fault detection and recovery (FDR) framework in autonomous robotic systems, which should be able to promptly detect  fault events and reliably adapt robot motions according to environmental changes. 

To address this, the robotics community has developed a range of FDR methods that aim to reduce fault detection latency and enhance fault recovery robustness using visual observations, which can be broadly classified into two categories: Planning Domain Definition Language (PDDL)-based method \cite{pddl,pddl2} and large model-based method \cite{10802284, 10801352, huang2024rekep}. \color{black}
PDDL-based methods model the robotic system as a finite state machine, where the system state is expected to evolve along predefined transitions based on discrete robotic motions. 
In this way, fault detection is performed by comparing the actual system state with the expected outcome, and recovery is achieved by restoring the system to a goal-satisfying state.
Although such state comparison enables fast fault detection, the recovery process requires extensive manual modelling of all motions and task rules.
To address this, recent works explore the use of vision language models (VLMs) and large language models (LLMs) for robotic FDR~\cite{10802284, 10801352, huang2024rekep} to improve both FDR timeliness and adaptability.
They use a LLM to generate and verify robot motions within constraints, thereby enabling fault detection and triggering fault recovery. 
The authors in \cite{10802284} generated and verified natural language constraints, such as ``robot must hold the item when manipulating it",  while applying prompt engineering to reduce the LLM inference time. 
In \cite{10801352}, VLMs are prompted to reason about whether the spatial relations between contactable parts (e.g., drawer knobs and robot fingers) satisfy  geometric constraints (e.g., fingers clamping door knobs), such as collinearity or perpendicularity.
A more direct approach was proposed in \cite{huang2024rekep} by extracting visual keypoints, and using their spatial proximity as both constraint and recovery goal for FDR.
However, although existing methods have reduced the FDR time through various mechanisms, a fundamental problem lies in their system design not being oriented towards the downstream task. In other words, their communication, computation, and control modules are not explicitly tailored to address the specific requirements of robotic FDR, leading to the following limitations:
\begin{enumerate}
    \item \textbf{Communication:} Robots typically lack onboard computational resource and must offload reasoning tasks to remote servers. In practice, existing frameworks have to perform periodical wireless communication of bandwidth-intensive visual data (e.g., images or point clouds) for task offloading, without considering whether such data is necessary. {For example, the RGB image used as input to the VLM is communicated with the robot at only 1 Hz in the real-world setup \cite{10802284};} 
    \item \textbf{Computation:} Their computing pipeline, from constraint generation, fault reasoning, to motion replanning, performs generic visual or language reasoning that contributes little to the FDR task, but results in unnecessary computation overhead and delayed decision-making. {For example, GPT-4-based robotic planning requires 15.38 s for inference, and even a fine-tuned lightweight OpenLLaMA-3B model still requires 2.87 s  \cite{10767280}, which is far above the sub-second regime required for fast robotic FDR;}
    \item \textbf{Control:} Their control strategy adopts a one-shot decision-making process without considering task-specific adaptation, which often fails to provide sufficient precision when precise control is required, while applying overly complex control when a minor positional adjustment is needed. {For example, external disturbances that change the object pose can prevent the robot from maintaining a stable grasp, thereby reducing the task success rate from 68.6\% to 46.7\% \cite{huang2024rekep}.}
\end{enumerate}

Motivated by the above limitations, we propose a Goal-oriented  Communication (GoC) framework for fast and robust robotic FDR, where we jointly design the communication-computing-control loop during the fault detection and recovery stages, with each module tightly coupled to achieve the goal of the downstream FDR task.
In the fault detection stage, inspired by the concept of GoC \cite{zhou2022task}, 
we innovatively define the 3D scene graph (3D-SG) as a more effective and explainable form of semantic representation for fault detection, and transmit the extracted 3D-SG only when a fault is detected instead of sending raw visual data periodically.
In the fault recovery stage, we train an expert Small Language Model (SLM) for robotic FDR 
to replace general-purpose PPDL replanners \cite{pddl,pddl2} or LLMs \cite{10802284, 10801352, huang2024rekep} used in prior works, enabling faster inference and more accurate generation of recovery motions.
Meanwhile, to improve control quality, we propose a task-adaptive two-way fault recovery module that adjusts the control precision according to the level of motion uncertainty. For faults that result in high positional deviation or require accurate physical interaction, we introduce a lightweight digital twin reconstruction module tailored for robotic FDR, which simulates and refines the recovery motions generated by the SLM.
Unlike conventional digital twins that reconstruct the entire scene geometry, our design adheres to the goal-oriented principle by reconstructing only task-relevant object contours, supported by a dedicated goal-oriented communication mechanism.
Our main contributions are summarised as follows:
\begin{itemize} 
\item[{\textbullet}] {\textbf{System-Level Contributions:} We propose a novel GoC framework for fast and robust robotic FDR, where the communication-computation-control loop follows a goal-oriented design that transmits only goal-oriented information, performs task-specific computation, and enables task-adaptive control, with the goal of minimising the FDR time while maximising the robotic task success rate.}
\item[{\textbullet}] {\textbf{Algorithm-Level Contributions:} We design a dual-branch semantic representation extraction module for both fault detection and digital twin reconstruction. For fault detection, we define the 3D-SG as the semantic representation, which is extracted by a Triplet Graph Convolutional Network (TripletGCN) and transmitted only upon fault occurrence. For digital twin reconstruction, we define object edge points as the semantic representation, which are extracted through attention-based edge point sampling and curve fitting, and transmitted when fine-grained control is required. For fault recovery, we develop an expert SLM tailored for generating recovery motions for robots, which is fine-tuned with Low-Rank Adaptation (LoRA) and refined through knowledge distillation from online LLM.
For robotic control, we propose a task-adaptive two-way control strategy that triggers the digital twin to verify and improve the SLM-generated recovery motions when necessary.}
\item[{\textbullet}] {\textbf{Experimental-Level Contributions:}} We validate the effectiveness of our GoC framework in Mujoco robotic simulations \cite{mujoco} over multiple representative robotic tasks. Simulation results show that our framework can effectively reduce the FDR time while achieving higher task success rate compared to the state-of-the-art (SOTA) frameworks.
\end{itemize}

The rest of this paper is organised as follows: Section \ref{sec:2} reviews related works.
Section \ref{sec:3} presents the system model and  formulated problem. 
Section \ref{sec:4} introduces different modules of our  GoC framework for FDR. 
In Section \ref{sec:sim}, simulation results are presented to verify the effectiveness of our proposed framework. 
Section \ref{sec:conclu} concludes the paper.

\section{Related Work} \label{sec:2}
In this section, we review the relevant works concerning goal-oriented communication, large language model, and digital twin reconstruction. We further discuss the key differences and novelty of our work compared to these existing studies.
\subsection{Goal-oriented Communication}
To address the communication challenges posed by bandwidth-intensive vision data, the concept of goal-oriented communication has emerged as a promising solution \cite{zhou2022task}.          
Different from traditional bit-oriented communication, it focuses on transmitting only the semantic representation of information needed to accomplish a specific communication goal, rather than the raw sensory data itself \cite{9955525}.

Recent studies have applied goal-oriented communication to vision tasks such as visual question answering \cite{sige}, metaverse reconstruction \cite{10577270}, and scene classification \cite{9796572}.
However, these studies are designed for specific tasks and
scenarios, therefore not directly applicable to robotic FDR.
This motivates us to rethink what data to transmit and when to transmit it for the robotic FDR task.
In this work, we innovatively define the 3D-SG as the semantic representation for goal-oriented communication in robot vision.
Unlike most existing works on image-based GoC that define the semantic representation as the semantic features\cite{9953076} or 2D-SG \cite{sige},
our 3D-SG encodes object attributes and spatial relationships in 3D space in a compact symbolic structure.
{This extends image-plane layouts (e.g., above or right) based on 2D projections into spatially 3D representations (e.g., standing on or next to), which provides depth estimation and geometric abstraction necessary for robotic FDR. Moreover, unlike other 3D semantic representations such as keypoints and 3D bounding boxes that mainly describe object locations or sparse geometric anchors, our 3D-SG serves as a higher-level abstraction for scene understanding by explicitly representing spatial relationships.}
Nonetheless, extracting 3D-SG from dense, noisy, and partial point clouds in real-world settings presents new technical challenges.

{Beyond task-specific applications,  goal-oriented communication has also been adopted to bridge sensing, communication, computation, and control.
An early \cite{R1} proposed a goal-oriented integration framework and defined a unified goal-oriented metric that measures the probability the system goal is achieved, which was further extended to address how sensor information should be prioritised for transmission and coordinated with control\cite{R2}.
In parallel, the authors of \cite{R3} introduced the Goal-oriented Tensor (GoT) to quantify how semantic mismatch arising from sensing and transmission leads to utility loss in control process, where GoT was further deployed to determine the optimal sampling and transmission frequencies, as well as control decisions \cite{R4}.
However, these existing studies abstracted the system as a Markov Decision Process, while sensing, transmission, and control are represented as state acquisition, update, and control, respectively.
An unsolved question remains how to move beyond this abstraction and instantiate goal-oriented communication for system-level co-design in real-world systems.
}


\subsection{Large Language Model}
The use of LLM has become increasingly prevalent in robotic task and motion planning \cite{roco, 10802322}. 
One of the earliest studies is the dialectic multi-robot collaboration framework \cite{roco}, where each robot is assigned its own LLM and discusses with each other to accomplish collaborative tasks.
This framework was further refined in \cite{10802322}, where sub-tasks are allocated to robot sub-teams based on  robot available skills, thereby seamlessly generalize to new environments.
However, despite these extensive efforts to enhance the reasoning capabilities of LLMs, a frequently overlooked challenge is the considerable inference time associated with their large model size, which prevents their deployment in real-time robotic systems. Although techniques such as stream interpreting\cite{Song_2023_ICCV} and prompt engineering \cite{10802284} have attempted to mitigate the latency, they remain constrained by the inherent computation overhead of large-scale architectures.
To this end, we propose to develop task-specific SLMs as a promising alternative. 
By sacrificing general reasoning capability to achieve more efficient inference, SLMs can be fine-tuned by domain-specific data to meet the timeliness requirement in robotic control.
{Nevertheless, current research on task-specific SLMs for robotics remains scarce, and mostly limited to less complex tasks such as mobile robot navigation \cite{10767280}, task decomposition \cite{choi2024llmsreasoningpotentialsmall}, and marine robots language-to-action mapping\cite{11128181}.
The two studies most relevant to our work are \cite{ravichandran_prism} and \cite{10.5555/3692070.3692417}, both of which applied distilled SLMs to perform task planning for robotic arms, with the former combining historical observation-action context to generate optimal actions, and the latter implementing in-context learning to abstract SLM decision-making as an observation and reward-driven action selection process similar to Reinforcement Learning.
However, the aforementioned works on SLM-based dynamic planning  mainly considered single-robot tasks in relatively stable environments, where the system state did not change abruptly and decision-making was typically sequential planning under low uncertainty, while our work focuses on fast FDR in highly dynamic and uncertain scenarios.}
More importantly, in our proposed GoC framework, the input to SLM is no longer raw vision data but compact semantic representation, which poses a new challenge in coupling the communication and computing.


\subsection{Digital Twin Reconstruction}
Digital twin technology has been widely deployed in industrial automation to monitor physical systems in real time~\cite{8477101}. 
Traditional digital twin approaches focus on high-fidelity reconstruction of the entire physical environment, with the aim of enabling precise robotic control in different scenarios, such as robot arm manipulation \cite{11016707}, UAV target searching \cite{10045049}, and mobile robot navigation \cite{10811853}.
However, one fundamental issue is that maintaining an accurate digital twin inevitably demands large communication and computational resources, which introduces significant transmission latency and rendering delay.
Previous works have made considerable efforts to reduce these delays through different methods, such as data compression\cite{10242296} and prediction \cite{9953092}.
Nevertheless, these approaches are still driven by the goal of reconstructing high-quality digital twins, aiming to recover raw data as accurately as possible.
In this work, we offer a new goal-oriented perspective, i.e.,  \textit{the reconstruction process should be guided by its utility for downstream tasks, rather than by the fidelity of raw data recovery.}
{Recent efforts have taken initial steps toward this perspective \cite{dai2024acdc, DT1, DT2, DT3}.
The recently proposed digital cousin \cite{dai2024acdc} replaces real-world objects with standardized 3D models that share similar geometric and semantic affordances to approximate digital twin. Further, digital genealogy was proposed to describe and generate objects through their inherent attributes such as shape and material \cite{DT1}. 
More relevantly, the authors of \cite{DT2} and \cite{DT3} developed digital twins for robotic arms and UAVs by adjusting the transmission frequency, interpolation strategy, and the prediction horizon, respectively. The former aimed to minimise the positioning error of the robotic arm rather than the accuracy of the reconstructed 3D model for remote teleoperated grasping, while the latter was designed to support collaborative UAV tracking.
However, these works still prioritise structural completeness over task specific utility. That is, although they have consciously distinguished what information is more useful than others, they still aimed to preserve the digital twin reconstruction quality rather than improving downstream task utility, with the potential of goal-oriented design unexploited.
}

\section{System Model and Problem Formulation} \label{sec:3}
In this section, we first introduce the real-time robotic FDR task. Based on this, we present the SOTA framework and our proposed GoC framework for fast and robust robotic FDR. Finally, we formulate the robotic FDR problem.

\subsection{Real-time Robotic FDR Task}
We consider a wireless robotic FDR system consisting of three components: 
(1) \textbf{the physical robotic workspace}, which includes multiple robot arms, target objects, and other  environmental fixtures (e.g., tables, pallets, and shelves);
(2) \textbf{the user equipment (UE)}, i.e., an RGB-D camera that captures the real-time scene images from a third-person view, and transmits images or extracted representation information to the edge server through a wireless link;
(3) \textbf{the edge server}, which runs LLMs or SLMs to generate recovery motions for the robot based on the received observations. 
During the task execution, the RGB-D camera \cite{camera} periodically captures an RGB image $\boldsymbol{R}_i  \in \mathbb{R}^{h\times w\times3}$ and a corresponding depth map $\boldsymbol{D}_ i \in \mathbb{R}^{h\times w}$ of the robotic workspace, where $i$ denotes the frame index, $h$ is the image height, and $w$ is the image width.
The robots collaboratively interact with physical objects by executing a sequence of predefined high-level motions to accomplish collaborative tasks, such as workpiece sorting and parcel palletising. 

{However, unexpected faults may occur during task execution, which requires a fast and robust FDR framework. 
In the robotics community, the control is commonly organised in a hierarchical manner, where high-level task planning determines what actions (e.g., ``pick" and ``place") should be executed and in what order, while low-level motion planning determines how each action should be physically executed through feasible and collision-free trajectories. 
Inspired by this, we categorise the faults that may arise during task execution into task-level faults and motion-level faults. This is because the detection and recovery of these two types of failures require different observation granularity, computation complexity, and control precision, and should therefore be handled differently to ensure that only the necessary communication, computation, and control are provided, without being excessive or insufficient.}
\begin{itemize}
    \item \textbf{Task-level Faults:} These faults refer to failures in accomplishing the intended high-level task goals, even when each motion is executed correctly. For example, the robot may place an object in the wrong bin or follow an incorrect placing sequence, even if the ``place" action succeeds. {Common task-level faults include stacking objects in an incorrect order, assigning a target object to the wrong target container, or repeating an already completed action.} Recovering from such faults  requires reasoning over the task context and replanning of subsequent actions to realign with the original task goals.
    \item \textbf{Motion-level Faults:} These faults refer to errors that occur during the execution of individual motions due to physical interaction uncertainties. For example, the robot may  collide with an obstacle due to partial observation, or fail to reach a target object because the object lies outside kinematic workspace. {Common motion-level faults include collision risks, infeasible reaching, and interference between robots in the shared workspace.} Recovering from such faults requires more precise and fine-grained control strategies, such as replanning robot trajectories at the waypoint level to ensure successful motion execution.
\end{itemize}

\subsection{State-of-the-art Fault Detection and Recovery Framework} \label{sec:trad}
\begin{figure*}
\centering
\includegraphics[width=0.79\linewidth]{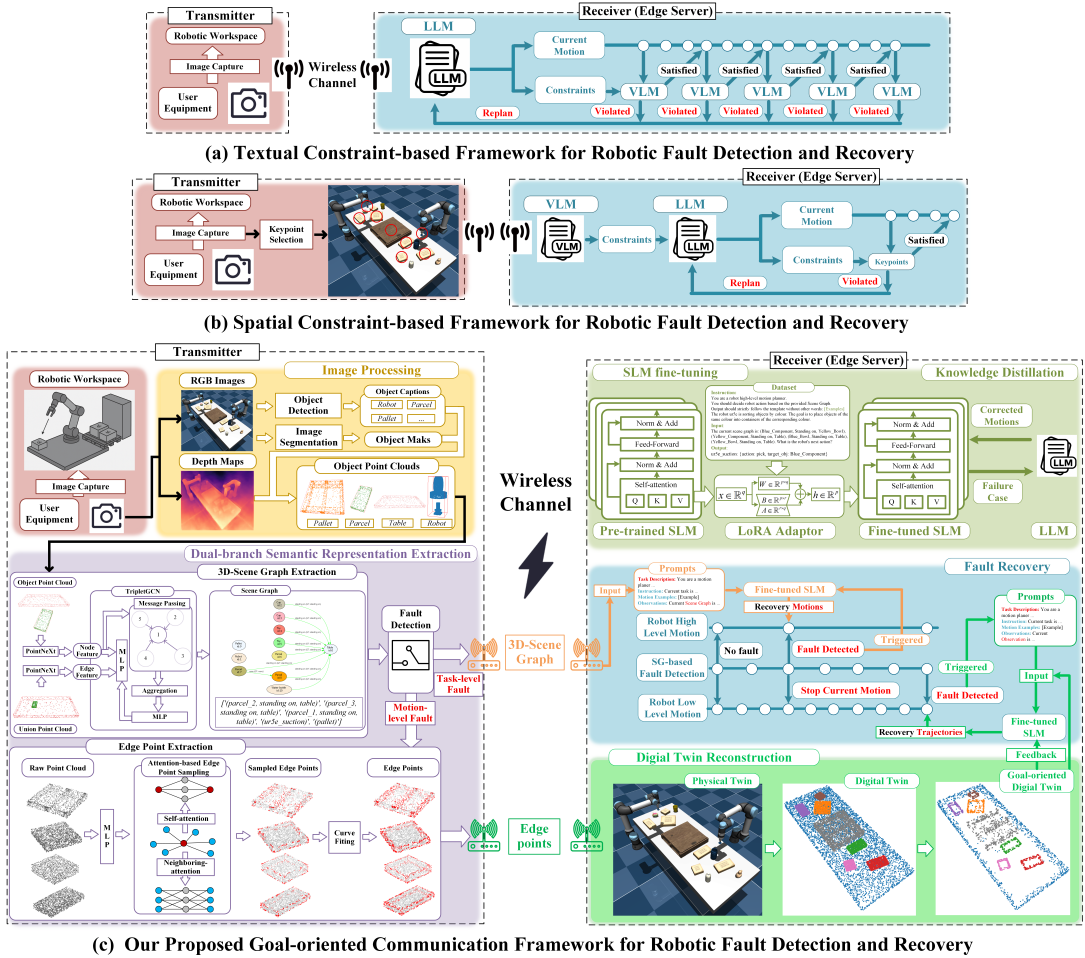}
\caption{State-of-the-art Frameworks and Our GoC Framework for Robotic Fault Detection and Recovery.}
\label{fig:sota}
\vspace{-0.3cm}
\end{figure*}

In this section, we review two representative SOTA frameworks that serve as baseline frameworks in this work, whose workflows are illustrated in Fig. \ref{fig:sota}(a) and  Fig. \ref{fig:sota}(b).
To adapt to diverse fault types and varying environments,
SOTA frameworks normally rely on large models to generate constraints for the robot's ongoing motion, and perform fault detection by periodically verifying whether these constraints are satisfied. The detected constraint violation will trigger a fault recovery mechanism that queries the LLM for corrective actions based on the latest observation and context. 

\subsubsection{Textual Constraint-based Frameworks}
These frameworks \cite{10802284}\cite{10990233} first leverage LLMs to generate explicit textual constraints corresponding to the robot's tasks and ongoing actions. For example,  consider the robot arm executing a ``move to table" motion after a ``pick up cup" motion, the LLM will automatically generate a set of constraints such as ``robot holding cup" and ``robot near table". The former typically corresponds to a motion-level constraint, which represents a necessary condition for the execution of the current motion. The latter corresponds to a task-level constraint, which identifies whether the current motion contributes to achieving the robotic task goal.
To verify whether such constraints are satisfied, the UE periodically transmits the captured image $\boldsymbol{R}_i$ to the edge server through the wireless link, where a VLM is deployed to perform spatial reasoning and assess whether constraints are satisfied in the current visual context. 
If the VLM determines that a constraint is violated, the edge server sends an emergency stop command to the robot and prompts the LLM to replan robotic motion using predetermined instructions.


\subsubsection{Spatial Constraint-based Frameworks}
Instead of transmitting images and using VLM to detect faults, these frameworks \cite{huang2024rekep} represent robotic tasks as a sequence of spatial constraints defined over spatial keypoints, which focuses on geometric relations such as contact, alignment, and proximity. For example, in the same ``move to table'' case, each object is abstracted as a single spatial keypoint (e.g., robot gripper, cup handle, table centre) and constraints are defined based on these keypoints, where the distance between the robot gripper and the cup handle serves as a motion-level constraint, while the distance between the robot and the table represents a task-level constraint.
Specifically, given the captured RGB-D images $\boldsymbol{R}_i$ and $\boldsymbol{D}_i$, the UE first performs keypoint selection locally. It applies DINOv2 \cite{dino} and SAM \cite{sam} to extract object masks and corresponding visual features, from which a small set of keypoint candidates are generated via $k$-means clustering. Redundant candidates within a fixed distance threshold are filtered out, and one representative keypoint is retained per object to serve as its spatial abstraction in the downstream constraint reasoning process. The resulting keypoints are then wirelessly transmitted to the edge server, where a VLM is prompted once before each robot motion with the RGB image overlaid with keypoints to generate a set of constraint functions. These constraint functions are expressed as executable programs and verified periodically without further VLM involvement. Similarly, if any constraint is violated, a LLM will intervene and replan the robot motions.





\vspace{-0.3cm}
\subsection{Our Goal-oriented Communication Framework} 
In this subsection, we present an overview of the key modules in our proposed GoC framework, as shown in Fig.~\ref{fig:sota}(c).
Different from the aforementioned SOTA frameworks, our GoC framework jointly designs the communication-computing-control loop, where all components ultimately serve the downstream task, i.e., fast and robust FDR.
Specially, we adopt a dual-branch architecture to address task-level and motion-level faults based on their distinct requirements for control granularity.
For task-level faults that require high-level geometric reasoning and symbolic planning, we extract and transmit a compact 3D-SG to the edge server, enabling fast fault detection and high-level replanning via the SLM.
For motion-level faults that demand fine-grained waypoint corrections for robots, we further extract and transmit object edge points at the UE side, and reconstruct a lightweight digital twin at the edge server to iteratively refine the recovery trajectories generated by SLM.


\textbf{Image Processing:} The captured RGB image $\boldsymbol{R}_i$ is first fed to YOLO11n-seg \cite{Yolo} to perform object detection and image segmentation, and generate a set of object representations $\mathcal{O}_i = \{O^j_i=(c_i^j, \boldsymbol{M}_i^j)\}_{j=1}^{N_i^O}$, where $c_i^j$ and $\boldsymbol{M}_i^j$ denote the caption and mask of the $j$-th object in frame $i$, respectively, and $N_i^O$ is the number of objects observed in frame $i$. For each object \(O^j_i\), we also extract its three-channel point cloud $\boldsymbol{P}_i^j$ through back-projecting the depth values within its mask \(\boldsymbol{M}^j_i\) using the depth camera intrinsic parameters, where we define the point cloud set of all objects as $\mathcal{P}_i = \{\boldsymbol{P}_i^1, \ldots, \boldsymbol{P}_i^{N_i^{O}} \}$.

\textbf{Dual-branch Semantic Representation Extraction:}
Instead of transmitting all the raw point cloud, our GoC framework extracts the semantic representation from them, to retain only necessary information required for downstream tasks, i.e., fault detection and digital twin reconstruction.


For fault detection, 3D-SG is defined as the semantic representation, which is formatted as an undirected graph composed of nodes and edges, where each node corresponds to an object and each edge describes a spatial relationship between two objects.
The object point cloud set $\mathcal{P}_i$ is fed into the representation extractor to generate the scene graph $\Phi_i$. 
For arbitrary object pair $(O^m, O^n)$, we use their captions and the undirected spatial relationship between them $ e_i^{m,n}$ to form a triplet $\phi = (c^m, e^{m,n}, c^n)$.
The 3D-SG for frame \(i\) is then constructed as a set of such triplets \(\Phi_i = \{\phi^1, \ldots, \phi^{N_i^{\Phi}} \}\), where \(N_i^{\Phi}\) is the number of valid object pairs in the 
 $i$-th frame. 
 
For digital twin reconstruction, object edge points are defined as the semantic representation, which correspond to a subset of the original point cloud that outlines the geometric contours of each object. We define the contour point subset of arbitrary object $O^m$ as $\boldsymbol{\hat{P}}_i^m$, and the
union set of contour points for all objects in the scene is then given by 
$\mathcal{\hat{P}}_i = \{\boldsymbol{\hat{P}}_i^1, \ldots, \boldsymbol{\hat{P}}_i^{N_i^{O}} \}$.

\textbf{Fault Detection:}
Based on the generated 3D-SG, fault detection is performed by capturing the geometric relational changes in the 3D environment. 
Specifically, we distinguish between task-level and motion-level faults by analysing the type of violated spatial relationships in the 3D-SG. 
A task-level fault typically occurs when the expected symbolic transition is missing in the 3D-SG after a motion execution. 
For example, in a parcel palletising task, after the robot executes a “pick up parcel” motion, the expected 3D-SG should evolve from $(\text{parcel}, \textit{standing on}, \text{table})$ to $(\text{parcel}, \textit{grasped by}, \text{robot})$. If the updated 3D-SG fails to include any relation between the robot and the parcel, this implies that the motion did not achieve the intended goal, and a task-level fault is detected.

In contrast, a motion-level fault is identified when an unexpected proximity relation emerges between the robot and a task-irrelevant object. For instance, in the same parcel palletising task, if the 3D-SG contains a relation such as $(\text{robot}, \textit{next to}, \text{human})$ during motion execution, this suggests that the robot is about to collide with the nearby human operator. Unlike task-level faults, motion-level faults arise from low-level spatial uncertainties and thus require fine-grained control adjustments.


\textbf{Uplink Transmission:}
To reduce the communication overhead, our proposed framework transmits only the goal-relevant semantic representation (i.e., 3D-SG for task-level fault, and edge points for motion-level fault) instead of full raw sensor observations. 
Note that the transmission is only triggered when the change in SG does not match the expected change, which means a fault is detected and the robot is commanded to stop immediately.
Moreover, we design an adaptive computation offloading scheme (see Sec. \ref{sec:offload}) to dynamically decide whether to extract the semantic representation locally at the UE or offload raw point cloud data to the edge server for remote processing based on the available bandwidth, to ensure low-latency fault detection under varying wireless conditions.

\textbf{Two-way Fault Recovery:}
%
To reduce the significant inference latency in SOTA framework, we train a task-specific SLM tailored for robotic fault recovery at the edge side. Upon receiving the latest 3D-SG or edge points, we adopt a two-way recovery strategy to address the varied control requirements of task-level and motion-level faults.
For task-level faults, the SLM generates symbolic high-level actions (e.g., \textit{place parcel on pallet}) based on the extracted 3D-SG. 
For motion-level faults that require precise corrections, the edge server reconstructs a lightweight digital twin using the received edge points to simulate the physical workspace. The initial recovery motion generated by the SLM is executed within this digital twin, where the system verifies whether the initial motion successfully avoids collisions and completes the intended action. If the verification fails, the feedback is used to re-prompt the SLM for motion refinement. This process is repeated until a valid and safe trajectory is obtained, which is then transmitted to the robot for execution.

\textbf{Motion Execution:} After receiving the refined recovery motions from the edge server, the robots either map the high-level symbolic actions  into executable trajectories (for task-level fault), or directly executes fine-grained waypoint sequences refined by the digital twin (for motion-level fault).


\vspace{-0.2cm}
\subsection{Channel Model}
{We model the wireless channel between the UE and edge server as Nakagami-$m$ fading with fading power gain $g$ to simulate industrial wireless settings as it can characterise a wide range of fading models (e.g., Rayleigh and Rician fading), which enables more robust simulations that can model  the dynamic small-scale fading caused by interference, obstacles, and movement within the industrial environment.}
The probability density function of the channel fading power gain $g$ is
\begin{align} 
	f_G \left ({g}\right) = \frac{g^{m-1}}{\Gamma(m)} \left(\frac{m}{\Omega}\right)^{m} e^{-{m\over\Omega}g},
\end{align}
where $\Gamma(m)$, $m$ and $\Omega$ are the Gamma function, shape parameter, and scale parameter, respectively.
We also assume packets experience in-factory Non-Line-of-Sight path loss \cite{3gpp38901}
\begin{align} 
	\text{PL} = 18.6 + 35.7\log_{10} (d) + 20\log_{10} (f_c),
\end{align}
where $d$ denotes the distance between UE and edge server, while $f_c$ is the carrier frequency.
The overall channel gain $h$ and the system signal-to-noise ratio
(SNR) are then derived as
\begin{align}
    h = \frac{\mathbb{E} \left[\vert g \vert^2 \right]}{10^{(\text{PL}/10)}}, \quad
    \text{SNR} = \frac{Ph}{\sigma^2},
\end{align}
where $P$ is the transmit power and $\sigma^2$ denotes the Gaussian white noise power. 
The transmission latency between the UE to the edge server is then given by
\begin{align}
    t^\text{com}=\frac{\Psi}{B \log_2 (1+\text{SNR})},
\end{align}
where $B$ is the channel bandwidth, and $\Psi$ is  the size of the transmitted data, which can be either the uplink 3D-SG string, the uplink edge points, or the downlink control command.

\subsection{Problem Formulation}
\begin{figure}
\centering
\includegraphics[width=1\linewidth]{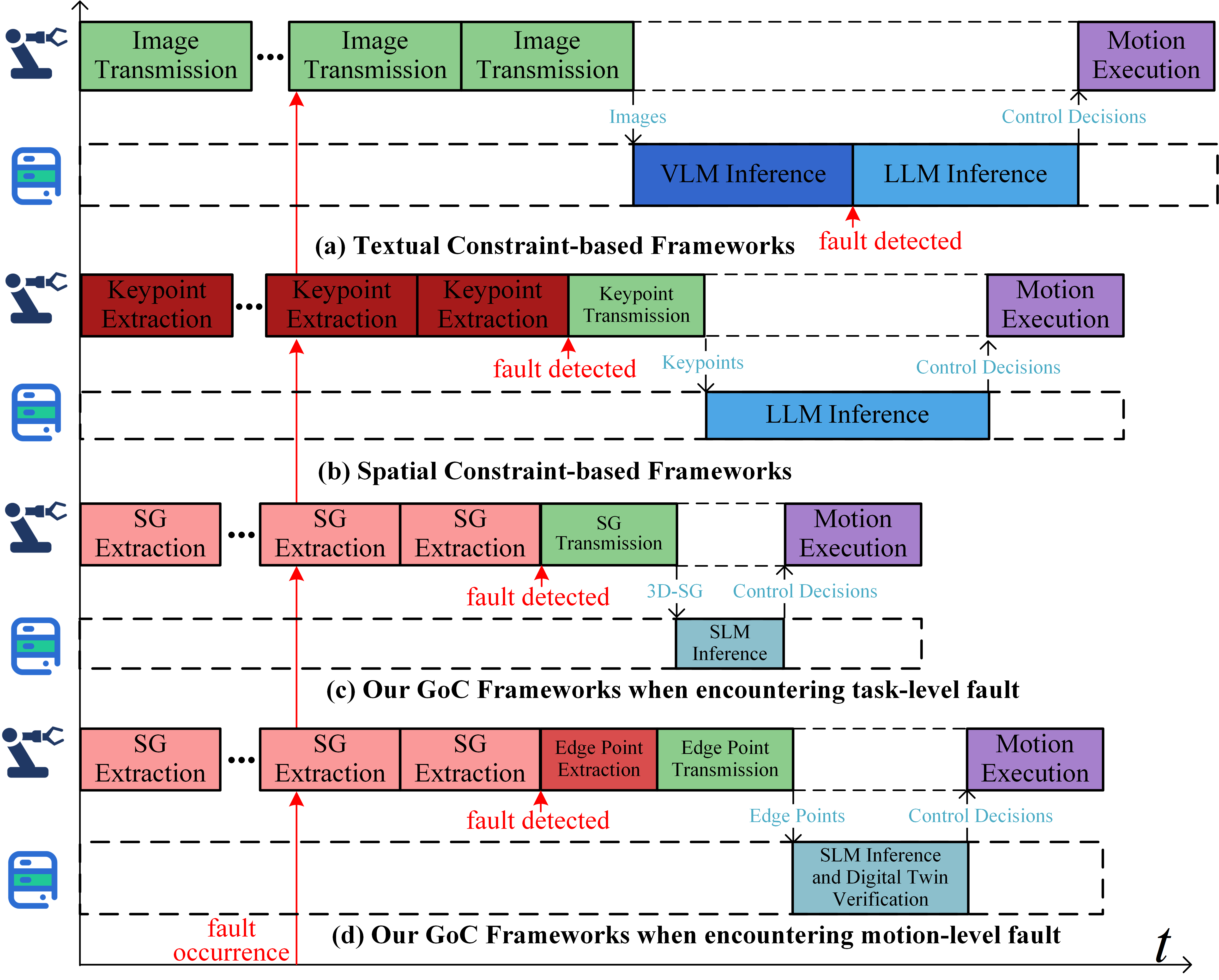}
\caption{The FDR time comparison.}
\label{fig:time}
\vspace{-0.3cm}
\end{figure}

As shown in Fig.~\ref{fig:time}, we define the FDR time $t^\text{FDR}$ as the time elapsed from the moment of fault occurrence in the robotic workspace to the completion of the recovery motion that allows the robot to resume its normal task execution. 
The FDR time of any task-level fault can then be expressed as
\begin{align}
    t^\text{FDR}=t^\text{sg} + t^\text{com} + t^\text{inf} + t^\text{exe},
\end{align}
where $t^\text{sg}$ and $t^\text{exe}$ denote the times for 3D-SG generation and recovery motion execution at the robot side, $t^\text{inf}$ denotes the SLM inference time at the edge server, and $t^\text{com}$ denotes the time for uplink 3D-SG transmission and downlink control command transmission. Also, when we perform additional edge point extraction and digital twin reconstruction to empower more precise control, the FDR time of motion-level fault is rewritten as 
\begin{align}
    t^\text{FDR}=t^\text{sg} + t^\text{dt} + t^\text{com} + t^\text{inf} + t^\text{exe},
\end{align}
where $t^\text{dt}$ denotes the edge point cloud extraction time at the robot side, while $t^\text{com}$ in this case denotes the transmission time for uplink edge point cloud and downlink control command. Note that here we omit the digital twin reconstruction and verification time, as these two processes are executed on the resource-abundant edge server and typically incur negligible computing latency (typically under 2ms).

In summary, the main \textbf{objective} of our proposed GoC framework is to minimise the overall robot FDR time $t^\text{FDR}$ while maximising the robotic task success rate.

\section{Our Proposed GoC Framework} \label{sec:4}
In this section, we detail the design of our proposed GoC framework. We first introduce the dual-branch representation extraction module, which includes the process of generating the 3D-SG from the object point cloud for fault detection, and also the extraction of edge points for digital twin reconstruction. 
We then develop an adaptive computation offloading scheme that decides under which conditions computation tasks should be offloaded, with the goal of adapting our framework to varying computation and communication resources in different industrial settings.
Finally, we present the fine-tuning process of the SLM for fault recovery.

\subsection{Scene Graph Generation}
Robotic fault detection fundamentally relies on understanding the spatial relationships among objects in the workspace.
To this end, we propose to use the more direct and efficient 3D-SG to represent the workspace spatial relationships and enable lightweight wireless transmission.
For all the objects  $\mathcal{O}_i = \{{O}_i^1, \dots , {O}_i^{N_i^O}  \}$ observed in image frame $i$, we first connect every pair of nodes with an undirected edge to form a fully-connected scene graph, which means a candidate spatial relationship initially exists between every pair of objects. We then filter these candidate edges by computing the spatial distance between the connected object pairs, and remove any edge whose distance exceeds a predefined threshold. This is because physically distant objects rarely have meaningful spatial relationships.

For each remaining object pair \((O_i^m, O_i^n)\) with a valid edge, we then predict their spatial relationship.
We first extract the individual node features $f^m_i$ and $f^n_i$ from their respective point clouds, and also the edge features $f^{m,n}_i$ from their merged point cloud. The node features encode the object geometric properties such as shape, orientation, and approximate category, while the edge feature captures their relative spatial configuration including contact and enclosure. All these features will be passed to the Triplet Graph Convolutional Network (TripletGCN)\cite{triplet} for spatial relationship prediction.
The motivation for using TripletGCN is its ability to capture global contextual information through iterative message passing over the graph, which allows each node to aggregate information from all other nodes. 
This global contextual information enables the model to develop a comprehensive understanding of the scene structure and enhance relationship prediction accuracy. 
For example, in a scene with a table, a book, and other small items, the model can more reliably infer the book is ``standing on" the table by incorporating global context such as the presence of all objects above the table.

\subsubsection{Node Feature and Edge Feature}
Each object's raw point cloud is first normalised by subtracting its centroid to ensure translation invariance, while deliberately avoiding global scaling in order to preserve the original spatial dimensions. 
The normalised point cloud is then uniformly downsampled to fixed size and passed to the PointNeXt \cite{pointnet} encoder to obtain the node features
\begin{align}
    f^m_i =  \text{PointNeXt} (\boldsymbol{\tilde P}^m_i), 
\end{align}  
where $\boldsymbol{\tilde P}^m_i$ denotes the normalized and downsampled point cloud of object $O_i^m$.

To obtain the edge features, we first concatenate the raw point clouds $\boldsymbol{P}_i^m$ and $\boldsymbol{P}_i^n$ of objects $O_i^m$ and $O_i^n$ into a union point cloud set $\boldsymbol{P}_i^{m,n}$. To retain point-object correspondence within the union set, we augment each point with an additional binary indicator channel $\delta$, where $\delta=0$ indicates the point is originally from object $O_i^m$, and $\delta=1$ otherwise. Similarly, the augmented point cloud is then normalised and passed through the PointNeXt encoder to obtain the edge feature $f^{m,n}_i$.

\subsubsection{Triplet Graph Convolutional Network}
Our TripletGCN consists of $N^L$ TripletGCN layers. For each layer $l$ and every valid object pair \((O^m, O^n)\)\footnote{We drop the subscript frame index $i$ for better explanation.}, given the node features $f^{m}$ and $f^{n}$, as well as the corresponding edge feature $f^{m,n}$, we form a feature triplet $f_l = (f^{m}_l, f^{n}_l, f^{m,n}_l)$ and input it to a multi-layer perceptron (MLP), which outputs updated feature triplet $f_{l+1} = (f^{m}_l, f^{n}_{l+1}, f^{m,n}_l)$. Then each node collects and averages the updated features corresponding to itself from all triplets it participates, and updates its own node feature as
\begin{align}
f^{m}_{l+1} = f^{m}_{l} + \text{MLP} \left( \frac{1}{|\Phi(m)|} \sum_{\phi \in \Phi(m)} f'^{m}_l(\phi) \right)
\end{align}
where \( \Phi(m) \) is the set of all other triplets that contain object \( O^m \), and \( f'^{m}_l(\phi) \) denotes the node feature of \( O^m \) in triplet \( \phi \), which can be obtained from the corresponding feature triplet. 
The output feature triplet \( f_{l+1} = (f^{m}_{l+1}, f^{n}_{l+1}, f^{m,n}_{l+1}) \) from layer \( l \) is then used as the input to layer \( l+1 \). 

After passing through \( N^L \) TripletGCN layers, we obtain the final edge feature \( f^{m,n}_{N^L} \) for each object pair \( (O^m, O^n) \) and feed it to a relationship classifier to predict the spatial relationship
\begin{align}
    P({e}^{m,n}) = \text{Softmax}(\text{MLP}(f^{m,n}_{N^L})),
\end{align}
where $ P({e}^{m,n})$ is the predicted probability distribution over all predefined spatial relationship categories of edge ${e}^{m,n}$.

To train the TripletGCN, we supervise the predicted relationship distributions using the ground-truth labels \( \bar{e}^{m,n} \) provided by the 3RScan dataset~\cite{3rscan}. The model is optimized end-to-end with the cross-entropy loss over all object pairs:
\begin{align}
\mathcal{L}_{\text{SG}} = - \sum_{(m,n)} \log P(e^{m,n} = \bar{e}^{m,n}).
\end{align}

\subsubsection{Sensitivity Analysis}
It is worth noting that our proposed fault detection module is relatively robust to perception errors in 3D-SG generation.
On the one hand, the impact of false positives depends on whether the error occurs at a task-critical relationship (i.e., a relationship contains the target object, the robot, and the target container) in the generated 3D-SG. If a false positive only occurs at task-irrelevant relationships, it will not affect  fault detection and downstream control decisions because the fault detection does not rely on perfectly recognising every spatial relationship in the scene, but mainly on whether the task-critical relationships evolve according to the expected task progression. In the worst cases where a task-critical relationship is incorrectly predicted, the main consequence is an unnecessary trigger of the fault recovery process, which increases the FDR time due to emergency stopping and replanning, but has limited impact on the final task success rate.
On the other hand, false negatives mainly delay fault detection because an actual fault is not recognised as soon as it occurs.  Nevertheless, this effect remains limited in our proposed framework because 3D-SG based fault detection is performed continuously, where the full 3D-SG generation pipeline, including image processing, point cloud construction, and relationship inference, is completed within approximately 150 ms. Therefore, in most cases, a fault missed in one cycle can still be captured in the next update, which constrains the propagation of perception errors to the downstream control logic.
\color{black}

\subsection{Edge Point Extraction}
At the robot side, once the 3D-SG detects that the robot is approaching an obstacle through the ``next to" relation, accurate motion adjustment becomes necessary to avoid collision. In such cases, geometric details of the environment are required to support collision-free path planning. To this end, we trigger the digital twin reconstruction module at the edge side for SLM to iteratively verify its generated motion plans.
However, a comprehensive digital twin typically encodes a large amount of redundant geometric details that are irrelevant to the downstream control task. To address this, we extract only the edge points of objects and transmit them to the edge server to reconstruct a lightweight digital twin. 
This is because path planning near obstacles primarily depends on the spatial boundaries of objects rather than their full surface geometry, which means edge points have provided sufficient information to identify collision margins and navigate through narrow spaces. 
In this subsection, we detail our edge point extraction module, where we first perform attention-based edge point sampling to capture object boundaries. Then, a curve fitting module reconstructs continuous object contours from the sparse edge points.

\subsubsection{Attention-based Edge Point Sampling}
Intuitively, points near object boundaries often lie at locations where the geometry changes abruptly, such as sharp edges and corners. To capture such geometric variability, we employ an attention-based sampling method \cite{10204840} that allows each point to evaluate how different it is from all others in the scene. In this way, edge points naturally receive stronger attention responses due to their structural distinctiveness.

For arbitrary point $p_t$ in the raw point cloud set $\boldsymbol{P}^m$ of object $O^m$, we first  encode it into a high-dimensional point feature vector \( f_t^m \in \mathbb{R}^{128} \) using a shared MLP, where we obtain a point feature set \(\boldsymbol{F}^m = \{f_t^m\}_{t=1}^{N_m}\) that captures the local geometric context of all points, in which \(N_m\) denotes the total number of raw points $\boldsymbol{P}^m$.

To capture the global structural relationship among all points, each feature \(f_t^m\) is then linearly projected into a query vector \(q_t^m \in \mathbb{R}^{64}\) and a key vector \(k_t^m \in \mathbb{R}^{64}\) via two independent linear layers
\begin{align}
    q_t^m = \boldsymbol{W}^Q f_t^m + b^Q, \quad
    k_t^m = \boldsymbol{W}^K f_t^m + b^K,
\end{align}
where \(\boldsymbol{W}^Q, \boldsymbol{W}^K \in \mathbb{R}^{64 \times 128}\) are weight matrices that linearly transform the 128-dimensional point features into the query and key embedding spaces, respectively, while \(b^Q, b^K \in \mathbb{R}^{64}\) are their corresponding bias. 
The attention score \(\alpha_{k,t}\) between point \(p_t\) and any other point \(p_k\) can then be computed by the dot product between their query and key embeddings
\begin{align}
    \alpha_{t,k} = \langle q_t^m, k_k^m \rangle, \label{equ:atten}
\end{align}
which allows each point to evaluate how similar or different it is with respect to all other points in the object.
To transform these similarity scores into a normalized attention distribution, we then apply a row-wise softmax operation, and obtain the normalised attention score, where Eq. (\ref{equ:atten}) can be rewritten as
\begin{align}
    \alpha_{t,k} = \frac{\exp(\alpha_{t,k})}{\sum_{k^{'}=1}^{N_m} \exp(\alpha_{t,k^{'}})},
\end{align}
in which the index \(k^{'}\) iterates over all points in the object \(O^m\). By computing normalised attention score \(\alpha_{t,k}\) for all point pairs \((p_t, p_k)\), we obtain a normalized attention matrix \(\boldsymbol{\alpha} \in \mathbb{R}^{N_m \times N_m}\), where each entry quantifies how much attention point \(p_t\) is assigned.

Finally, we compute the column-wise summation of the normalized attention matrix to measure the total attention each point receives, where the summed attention of point $p_t$ is
\begin{align}
    s_t = \sum_{k=1}^{N_m} \alpha_{k,t}.
\end{align}
Herein, points with higher attention scores are normally structurally distinctive to other points, which correspond to boundary or edge regions.
Therefore, we select the top \(k\) (512 in our cases) points with the highest values as edge points
\begin{align}
    \boldsymbol{\hat{P}}^m = \text{TopK}(\{s_t\}_{t=1}^{N_m}).
\end{align}

\subsubsection{Curve Fitting}
To  compensate for the incompleteness and sparsity of the edge points by producing a smooth and continuous approximation of the object boundary, we further reconstruct the contour of each object via curve fitting after extracting sparse edge points from the raw point cloud.

Given the sampled edge point set $\boldsymbol{\hat{P}}^m$ for object \(O^m\), we first project these points onto three 2D planes, such that the primary geometric variation is preserved in the projected coordinates, where $\boldsymbol{\hat{P}}^m_x$, $\boldsymbol{\hat{P}}^m_y$, and $\boldsymbol{\hat{P}}^m_z$  denote the resulting 2D edge point sets.
To avoid fitting across disjoint or structurally unrelated edges, we apply spatial clustering on the projected point sets to segment the edge points into locally continuous fragments.
Each fragment is then independently fitted into a parametric curve $C(t)$ using B-spline regression, where $t$ denotes pseudo-arc-length parameter. The fitted curve is regularized to ensure smoothness while minimising deviation from the original points through
\begin{align}
    \min_{C(t)} \sum_{g=1}^{G} \left\| (x_g, y_g) - C(t_g) \right\|^2 + \lambda \int \left\| \frac{d^2 C(t)}{dt^2} \right\|^2 dt,
\end{align}
where $G$ denotes the total number of points in current cluster, and $g \in \{1, \ldots, G\}$ is the index of each point. $(x_g, y_g)$ represent the 2D coordinates of the $g$-th point, \(t_g\) is pseudo-arc-length parameter, and \(\lambda\) is the regularization coefficient that controls smoothness.

\subsection{Adaptive Computation Offloading Scheme} \label{sec:offload}
To ensure low-latency fault detection under varying channel conditions and task requirements, our GoC framework integrates an adaptive offloading scheme that determines whether to perform the semantic representation extraction locally on the UE or offload the entire raw point cloud data to the edge server for remote processing. The core motivation is to minimise the fault detection time by adaptively selecting the timing and content of data transmission based on bandwidth availability. Specifically, the fault detection time of local computing  can be expressed as
\begin{align}
	t^\text{det}_\text{local}=\left\{
	\begin{array}{ll}
		t^\text{sg}_\text{local} + \dfrac{\Psi^{\text{sg}}}{B \log_2 (1+\text{SNR})}, & \text{task-level fault}, \\[6pt]
		t^\text{sg}_\text{local} + t^\text{dt}_\text{local} + \dfrac{\Psi^{\text{edge points}}}{B \log_2 (1+\text{SNR})}, \!\!\!\!\! & \text{motion-level fault},\\
	\end{array} \right.
\end{align}
where  $\Psi^{\text{sg}}$ and $\Psi^{\text{edge points}}$ denote the  data size of transmitted 3D-SG and edge points, respectively.
It can be observed that the detection time for motion-level faults is always longer than that for task-level faults, due to the additional digital twin reconstruction and the larger data size of edge points.

Also, the fault detection time under edge computing, where the entire raw point cloud is transmitted to the edge server for remote processing, can be expressed as
\begin{align}
t^\text{det}_\text{edge} = \dfrac{\Psi^{\text{full points}}}{B \log_2 (1+\text{SNR})} + t^\text{sg}_\text{edge} + t^\text{dt}_\text{edge},
\end{align}
where $\Psi^{\text{full points}}$ denotes the size of entire scene point clouds.

\begin{algorithm}[t]
	\caption{Adaptive Computation Offloading Scheme}\label{alg:1}
	\textbf{Input:} Real-time bandwidth $B$, local computation time $t^\text{sg}_\text{local}$ and $t^\text{dt}_\text{local}$, edge computation time $t^\text{sg}_\text{edge}$ and $t^\text{dt}_\text{edge}$\\
	\textbf{Output:} Offloading Choices
	\begin{algorithmic}
            \State Calculate bandwidth threshold $B_1$ and $B_2$, 
		\For{$i = 1$ to $N_T$} 
            \If{$B \leq B_1$}
	    \State Local computing for all faults
		\ElsIf {$B_1 < B \leq B_2$}
            \State Local computing for task-level faults, edge computing for motion-level faults
            \Else
            \State Edge computing for all faults
		\EndIf
		\EndFor
	\end{algorithmic}
\end{algorithm}

Since the computation time (i.e., $t^\text{sg}_\text{local}$, $t^\text{dt}_\text{local}$, $t^\text{sg}_\text{edge}$ and $t^\text{dt}_\text{edge}$) is primarily determined by fixed computation resource such as onboard processors and edge server capacity \footnote{{In practice, these computation times can be estimated in advance either by quantifying the workload using metrics such as FLOPs \cite{11051063} or CPU/GPU cycles \cite{10933516} and then calculating the corresponding execution time based on the throughput of the hardware, or by benchmarking on the hardware over several preliminary runs and recording the average execution time.}}, it remains invariant under changing bandwidth conditions. In contrast, the transmission time is highly sensitive to bandwidth availability, which becomes the dominant factor affecting the total fault detection latency.
Based on this, we aim to identify two critical bandwidth thresholds, denoted as $B_1$ and $B_2$ ($B_1 < B_2$), at which the total fault detection time of local computing becomes equal to that of edge computing, i.e., $t^\text{det}_\text{local} = t^\text{det}_\text{edge}$. 
In practice,  once the computing capabilities of the robot and the edge server are determined, the thresholds $B_1$ and $B_2$ can be obtained directly, since local and edge processing times $t^\text{sg}_\text{local}$, $t^\text{dt}_\text{local}$, $t^\text{sg}_\text{edge}$ and $t^\text{dt}_\text{edge}$ can be measured offline, while the data sizes $\Psi^{\text{sg}}$, $\Psi^{\text{edge points}}$, and $\Psi^{\text{full points}}$ are known at the transmitter before transmission.
Specifically,  the two thresholds serve as decision boundaries for selecting the optimal computation strategy under different bandwidth conditions:
\begin{align}
	\left\{
	\begin{array}{lll}
		\text{Local Computing}, & B \leq B_1, \\[6pt]
		\text{Hybrid Strategy}, & B_1 < B \leq B_2,\\[6pt]
            \text{Edge Computing}, & B_2 < B.
	\end{array} \right.
\end{align}
When the available bandwidth is below $B_1$,  transmitting raw point cloud is extremely time-consuming. In this case, the end-to-end latency is dominated by the transmission delay,  local computing with minimal semantic representation achieves lower fault detection latency and should be selected for any type of faults.
When bandwidth lies between $B_1$ and $B_2$, the reduction in transmission time brought by transmitting 3D-SG can still offset the lower computing latency at edge server. However, for motion-level faults that require more bandwidth-intensive edge points, the edge server’s superior computing power starts to compensate for transmission delay, which means it is preferable to adopt a hybrid offloading strategy that performs local computing for task-level faults and delegates motion-level fault processing to the edge server.
When bandwidth exceeds $B_2$, the enhanced channel capacity makes raw point cloud transmission feasible in real-time, thereby  offloading all computation tasks is the most time-efficient choice.

Finally, this offloading strategy is summarised in \textbf{Algorithm \ref{alg:1}}, which dynamically selects optimal computing location and transmission content to minimise the fault detection latency.

\subsection{Small Language Model Fine-tuning} \label{sec:slm}
Another major bottleneck of the SOTA framework lies in the fault recovery stage, where the use of LLM to generate recovery motions introduces considerable inference latency. 
However, we notice that  
1) the capability of LLMs is unnecessarily expensive for the fault recovery task with limited linguistic complexity, a fine-tuned expert SLM can already achieve comparable performance with reduced inference time;
and 2) the fine-tuning cost of the SLM is relatively low, which allows rapid deployment and adaptation to various robotic applications. 
Therefore, we replace the LLM with a fine-tuned SLM to further reduce the FDR time.

In this subsection, we present the pipeline of SLM fine-tuning, which consists of a LoRA fine-tuning module and a knowledge distillation module.
We fine-tune a pre-trained Llama-3.2-1B-Instruct model \cite{llama} for task-level fault recovery, and a pre-trained Llama-3.2-11B-Instruct for motion-level fault recovery, as extremely small models (e.g., models with 1B or 3B parameters) tend to struggle with numerically grounded reasoning and often fail to produce valid waypoints.

\subsubsection{LoRA Fine-tuning} 
Given the robotic task description, recovery motion examples, and the latest scene observation (i.e., 3D-SG for task-level fault or edge points for motion-level fault), we expect the SLM to generate recovery motions (i.e., task reallocation for task-level fault or valid waypoints for motion-level fault) in a predefined structured format.

\begin{figure}
\centering
\includegraphics[width=1\linewidth]{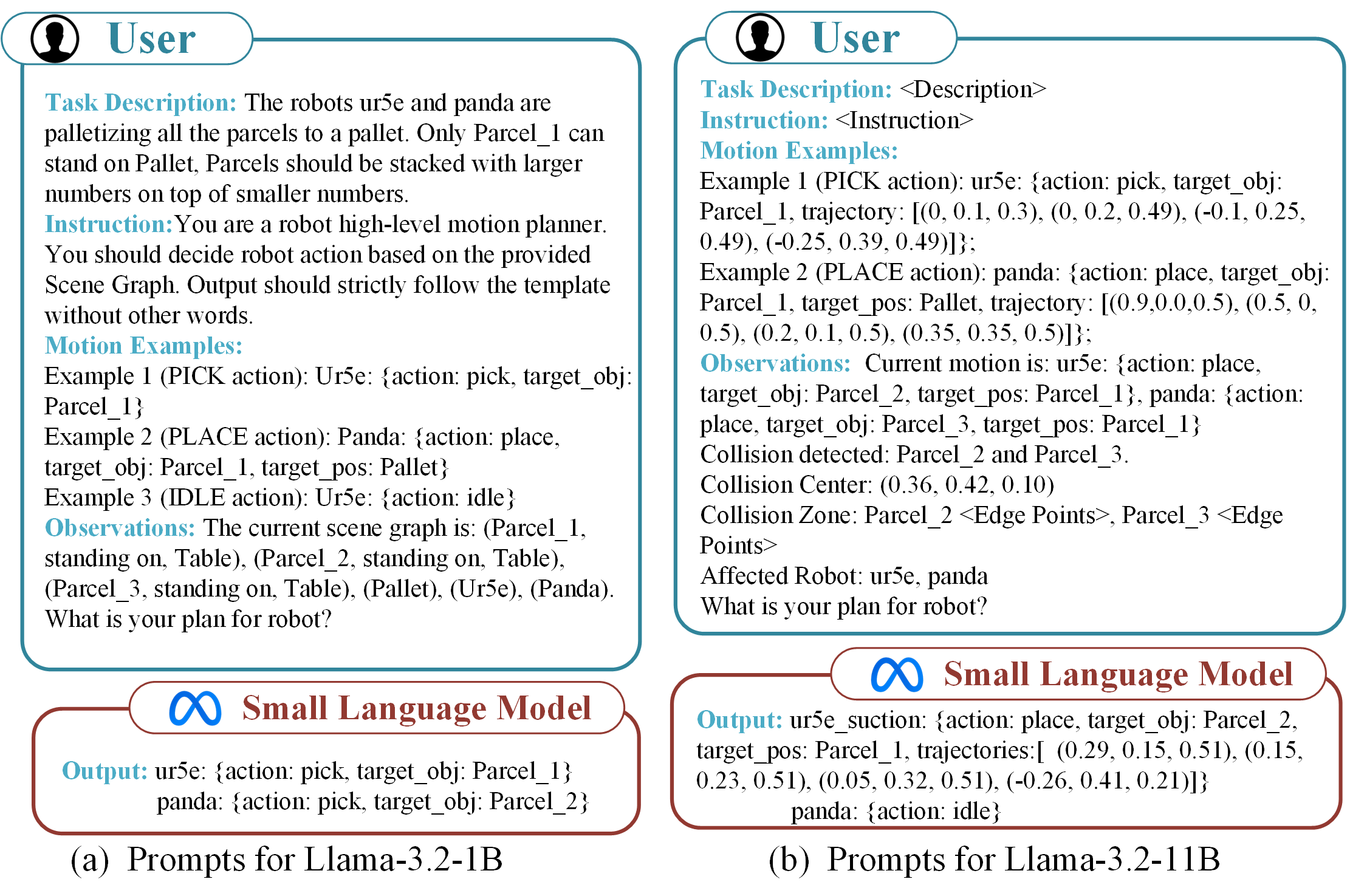}
\caption{Fine-tuning prompt example.}
\label{fig:prompts}
\vspace{-0.3cm}
\end{figure}

To construct the fine-tuning dataset, we first collect data by running robotic tasks simulations online and prompting GPT-4o to generate recovery motions based on observed failures. The generated samples are structured in Alpaca format, and then manually refined to ensure quality and consistency, where the example entry is presented in Fig. \ref{fig:prompts}. 

The fine-tuning objective is to align the output of the SLM with the ground-truth recovery motion. Let \( \mathcal{H} = \{(x_h, y_h)\}_{h=1}^{N^H} \) denote the constructed fine-tuning dataset, where \( x_h \) is the tokenised prompt constructed from task and scene information, and \( y_h = (y_{h,1}, \dots, y_{h,T_h}) \) is the corresponding recovery motion token sequence of length \( T_h \). 
The loss function is computed as the negative log-likelihood of generating the correct token
\begin{align}
\mathcal{L}_{\text{SLM}} = \frac{1}{N^H} \sum_{h=1}^{N^H} \left( - \frac{1}{T_h} \sum_{t=1}^{T_h} \log P(y_{h,t} \mid y_{h,<t}, x_h; \boldsymbol{\theta}) \right)
\end{align}
where \( P(y_{h,t} \mid y_{h,<t}, x_h) \) denotes the probability of generating the ground-truth token \( y_{h,t} \), conditioned on the prompt \( x_h \) and the previously generated tokens \( y_{h,<t} \), while $\boldsymbol{\theta}$ denotes the trainable parameters of the SLM.

During fine-tuning, we apply LoRA into the self-attention layers of the pre-trained SLM to accelerate fine-tuning. Specifically, we inject LoRA adapters into the query, key, and value projection matrices $\boldsymbol{W}^Q$, $\boldsymbol{W}^K$, and $\boldsymbol{W}^V$, while keeping the rest of the SLM model frozen. 
This is motivated by the observation that over-parametrised language model can be fine-tuned effectively within a low-dimensional intrinsic subspace of the original full parameter space without compromising the reasoning performance \cite{lora}.
Let us take the fine-tuning of query projection matrix $\boldsymbol{W}^Q_0 \in \mathbb{R}^{p \times q}$ as an example. Instead of updating $\boldsymbol{W}^Q_0$ based on gradients, LoRA introduces a trainable, additive weight update $\Delta \boldsymbol{W}^Q$, and decomposes it into pairwise low-rank matrices $\boldsymbol{A} \in \mathbb{R}^{r \times q}$ and $\boldsymbol{B} \in \mathbb{R}^{p \times r}$, such that query projection matrix updating becomes
\begin{align}
\boldsymbol{W}^Q_0 + \frac{r}{\alpha} \Delta \boldsymbol{W}^Q = \boldsymbol{W}^Q_0 + \frac{r}{\alpha} \boldsymbol{B} \boldsymbol{A},
\end{align}
with $r \ll \min(p, q)$ representing LoRA rank and $\alpha$ denoting the low-rank update magnitude. 
In this way, the number of trainable parameters drops from $p \times q$ to $r(p + q)$ without affecting the fine-tuning quality.

\subsubsection{Knowledge Distillation}
Although the fine-tuned SLM performs well in most cases, its reasoning capability is inherently limited by model size, which makes it less reliable when encountering complicated fault cases or unseen fault scenarios. 
For example, in parcel palletising task, suppose a parcel has been placed on the pallet and the robot is about to stack another on top. If an unexpected obstacle is left on top of the existing parcel, the SLM may fail to reason that the obstacle must be removed first before proceeding. 
To address this, we design a knowledge distillation module that enables the SLM to learn recovery strategies from a more knowledgeable LLM.

We adopt an online knowledge distillation design, where a teacher model (GPT-4o in our case) is only triggered when the SLM fails to generate a valid recovery motion\footnote{{We regard a recovery motion as valid only if it satisfies the predefined structured output format and successfully resolves the fault. Specifically, we validate each recovery motion generated by the SLM through executing it in the simulator and checking whether it leads to task progress, such as re-grasping an accidentally dropped object or avoiding a potential collision. If verification fails, the simulator restores the scene to the state before motion execution and triggers the teacher model for knowledge distillation.}}. At this time, we prompt the teacher model with the same scene observation, task description, and the failed response from the SLM to obtain a corrected recovery motion, accompanied by an explanation that analyses why the original reasoning of SLM was incorrect. 
Let $x$ denote the tokenised input prompt, and $y^{\text{LLM}}$ the recovery motions generated by the teacher model. We expect the SLM to imitate the output of teacher model and leverage the guidance to improve fault recovery quality.
The loss function of knowledge distillation is then defined as token-level cross-entropy loss between the outputs of SLM and teacher model
\begin{align}
    \mathcal{L}_{\text{KD}} = \mathbb{E}_{x \in \mathcal{X}_{\text{fail}}} \left[ - \frac{1}{T} \sum_{t=1}^{T} \log P (y_t^{\text{LLM}} \mid y_{<t}^{\text{LLM}}, x; \boldsymbol{\theta}) \right],
\end{align}
where \( \mathcal{X}_{\text{fail}} \) is the set of inputs when LLM is triggered, and $T$ is the token length of the LLM's output $y^{\text{LLM}}$. The loss function formulation follows the classical teacher forcing strategy to stabilise the knowledge distillation, where the ground-truth token prefix generated by the teacher model is used as input to the SLM at each step. In this way, the SLM can avoid predicting the next token using its previously predicted tokens as input, as they might already be incorrect. 
{Also note that our distillation is hard-labelled rather than soft-labelled. This is because our task requires structured recovery outputs with a fixed format, where precisely reproducing the correct output sequence is already sufficient, while requiring the student model to fully match the teacher's output distribution offers limited performance improvement.}

\section{Simulation Results}  \label{sec:sim}
In this section, we validate the effectiveness of our proposed GoC framework, and compare it with the SOTA framework, where the channel parameters, TripletGCN hyperparameters, and LoRA fine-tuning hyperparameters are listed in Tab.~\ref{tab:1}. 
{Note that the Nakagami-$m$ fading model is simplified to Rayleigh fading for simulating a general factory scenario with non-line-of-sight propagation and rich scattering \cite{nakagami}.}
\begin{table}[t]
\centering
\setlength{\abovecaptionskip}{-0.01cm} 
\caption{Simulation Setup.}
\renewcommand\arraystretch{1}
\label{tab:1}
\begin{tabular}{|cccc|}
\hline
\multicolumn{4}{|c|}{\textbf{Channel Parameters}} \\ \hline
\multicolumn{1}{|c|}{Fading Shape Parameter} & \multicolumn{1}{c|}{1} & \multicolumn{1}{c|}{Carrier Frequency} & 3.5GHz \\ \hline
\multicolumn{1}{|c|}{Fading Scale Parameter} & \multicolumn{1}{c|}{1} & \multicolumn{1}{c|}{Distance} & 50m \\ \hline
\multicolumn{1}{|c|}{Transmit Power} & \multicolumn{1}{c|}{{24dBm}} & \multicolumn{1}{c|}{Noise Power} & -114dBm \\ \hline
\multicolumn{1}{|c|}{Bandwidth} & \multicolumn{1}{c|}{{1MHz}} & \multicolumn{1}{c|}{RGB Image Size} & 1920×1080 \\ \hline
\multicolumn{4}{|c|}{\textbf{TripletGCN Hyperparameters}} \\ \hline
\multicolumn{1}{|c|}{Node Feature Size} & \multicolumn{1}{c|}{256} & \multicolumn{1}{c|}{Weight Decay} & 0.01 \\ \hline
\multicolumn{1}{|c|}{Edge Feature Size} & \multicolumn{1}{c|}{512} & \multicolumn{1}{c|}{Learning Rate} & 0.0001 \\ \hline
\multicolumn{4}{|c|}{\textbf{LoRA Hyperparameters}} \\ \hline
\multicolumn{1}{|c|}{LoRA Dropout} & \multicolumn{1}{c|}{0.1} & \multicolumn{1}{c|}{LoRA Bias} & None \\ \hline
\multicolumn{1}{|c|}{LoRA Scaling Factor} & \multicolumn{1}{c|}{16} & \multicolumn{1}{c|}{Weight Decay} & \multicolumn{1}{c|}{0.01} \\ \hline
\multicolumn{1}{|c|}{Optimizer} & \multicolumn{1}{c|}{Adam} & \multicolumn{1}{c|}{Learning Rate} & 0.001 \\ \hline
\end{tabular}
\vspace{-0.3cm}
\end{table}

\vspace{-0.3cm}
\subsection{System Setup and Baselines}
{We perform all robotic simulations using the MuJoCo simulator\cite{mujoco}, where we design and implement three representative industrial robotic tasks with increasing levels of difficulty:}
{
\begin{itemize}
    \item \textbf{Workpiece Sorting (Easy):} This task requires a single robot to sort three workpieces with different colours by placing each of them into the container with the corresponding colour. Accordingly, task success is defined as successfully sorting all three workpieces without being misplaced or unsorted.
    \item \textbf{Grocery Packing (Medium):} This task requires two robots to pick up grocery items (e.g., milk cartons, bananas, and cup) with different shapes and heights from a table and place them into a bin, without any requirement on the item order or exact placement position. Accordingly, task success is defined as successfully placing all target grocery items into the bin with a feasible space allocation.
    \item \textbf{Parcel Palletising (Difficult):} This task requires two robots to sequentially pick up parcels from a table and place them onto a pallet by following predefined placement orders, where smaller parcels are stacked on top of larger ones. Accordingly, task success is defined as successfully palletising all parcels in correct order while ensuring stable stacking.
\end{itemize}}

We also introduce random fault events to simulate manipulation noise in real-world industrial scenarios \cite{9505314}, including
\textbf{Unexpected Object Drops (Task-level):} The robot may accidentally release an object during manipulation, reflecting typical execution failures due to unstable grasps or actuation errors in real-world systems;
\textbf{Placement Noise (Task-level):} The object may be placed at an unintended position that deviates from the target location;
\textbf{External Disturbances (Motion-level):} A human operator may intentionally remove objects or deliberately obstruct the robot’s path, forcing the robot to fail to complete the intended action.

We compare our GoC framework with the following two SOTA frameworks that are introduced in Sec. \ref{sec:trad}:
\begin{itemize}
    \item \textbf{Textual Constraint-based Framework:} This framework transmits RGB image for visual input, where fault detection is performed via the VLM, and recovery motions are generated by the LLM.
    \item \textbf{Spatial Constraint-based Framework:} This framework transmits object keypoints for fault detection, and also uses the LLM for recovery motion generation. 
    \item \textbf{Our GoC Framework:} Our framework transmits 3D-SG for fault detection, object edge points for digital twin reconstruction, and uses an expert SLM for fault recovery. 
\end{itemize}
To ensure a fair comparison, all frameworks adopt open-source Llama-3.2 models \cite{llama}. Both SOTA frameworks employ the Llama-3.2-90B model for vision reasoning, constraint generation and replanning. Our GoC framework adopts a lightweight Llama-3.2-1B model as the task-level replanner, and a more capable Llama-3.2-11B model for motion-level replanning.


We evaluate 25 runs for each task and report the average FDR time and the overall task success rate to demonstrate the time-efficiency and the reliability of our proposed GoC framework.
The UE processes the images and performs representation extraction with a single NVIDIA 2060 GPU, while the edge server is equipped with four NVIDIA A100 GPUs to run the language models.
Also note that the FDR time is computed based on the number of simulation steps between the moment a fault occurs and the completion of the corresponding recovery motion for fair comparison.

\begin{figure*}[t]
    \centering
    \begin{subfigure}[b]{0.32\linewidth}
        \includegraphics[width=\linewidth]{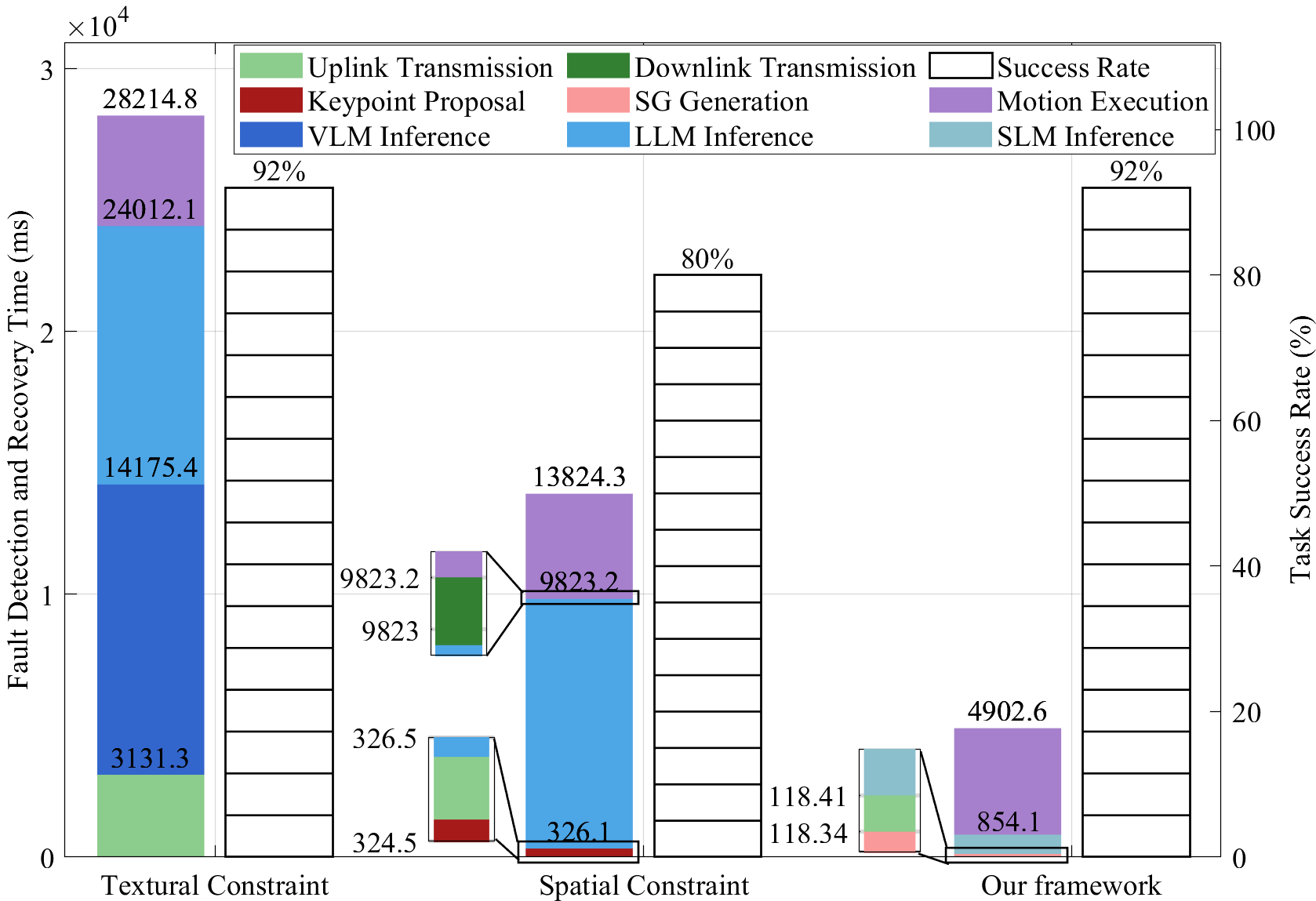}
        \caption{Workpiece Sorting.}
        \label{fig:s_place_load}
    \end{subfigure}
    \begin{subfigure}[b]{0.32\linewidth}
        \includegraphics[width=\linewidth]{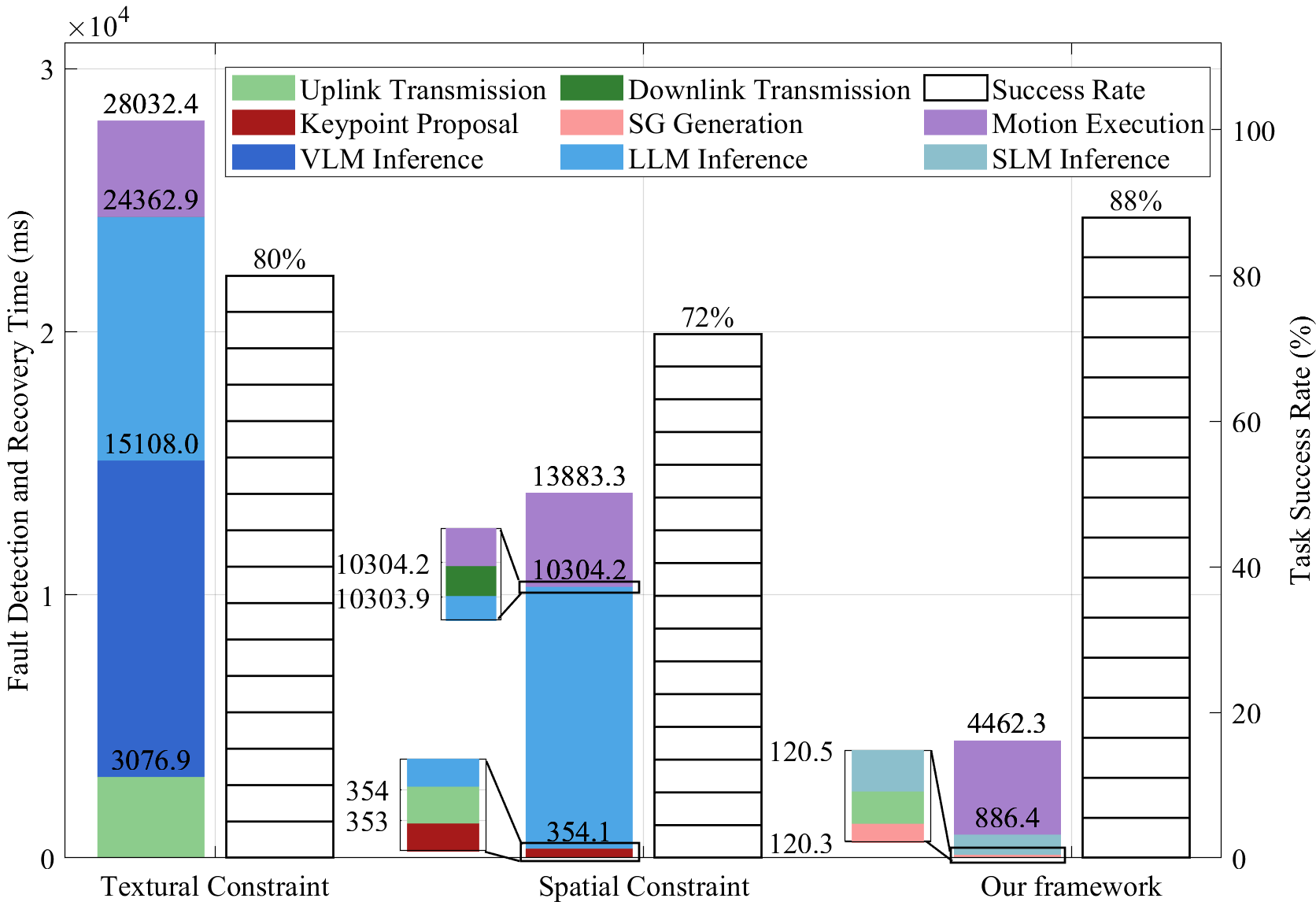}
        \caption{Grocery Packing.}
        \label{fig:s_place_angle}
    \end{subfigure}
    \begin{subfigure}[b]{0.32\linewidth}
        \includegraphics[width=\linewidth]{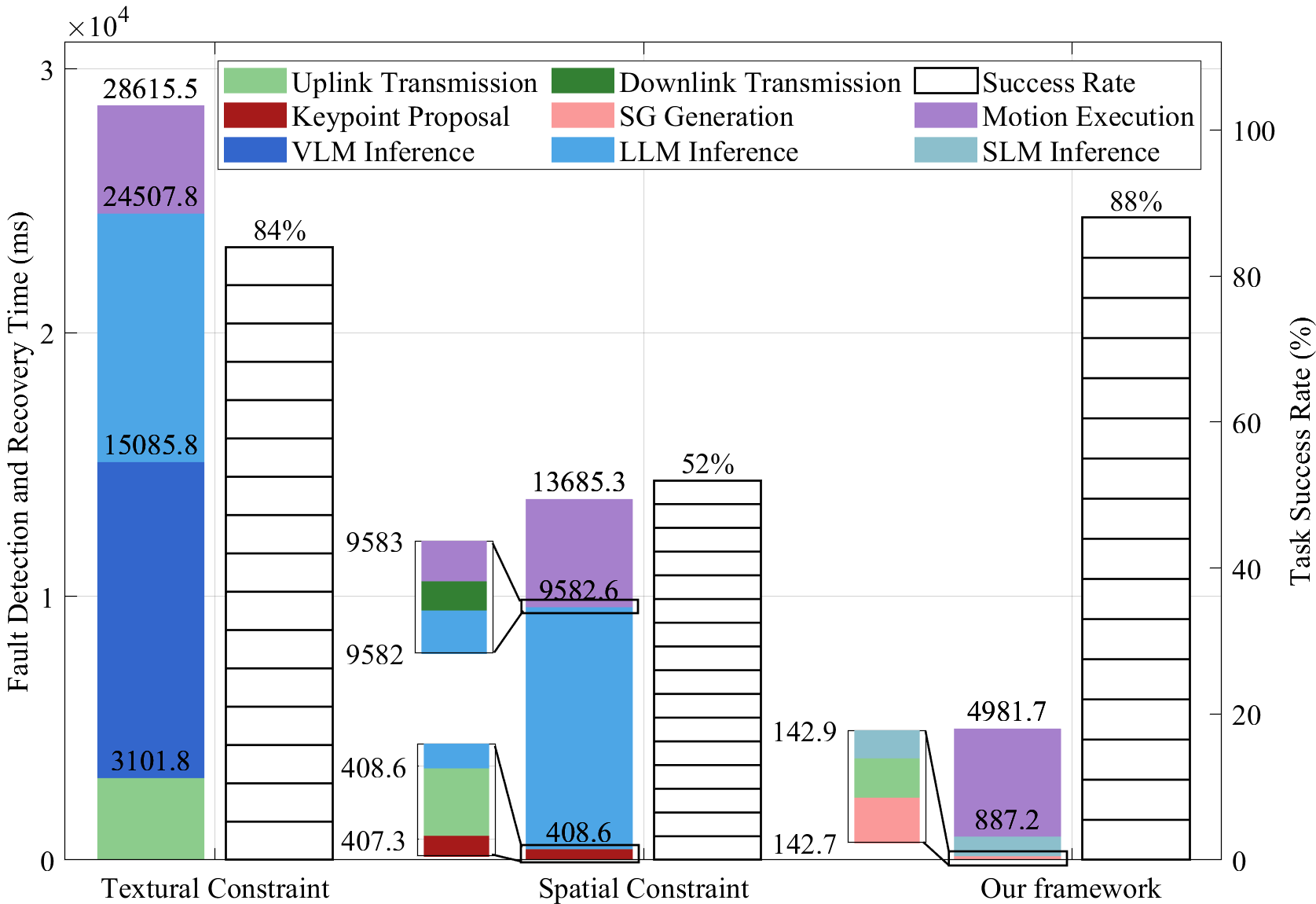}
        \caption{Parcel Palletising.}
        \label{fig:s_place_vel}
    \end{subfigure}
    \caption{\small Comparison of FDR time and task success rate between SOTA and our proposed GoC frameworks in handling task-level faults.}
    \label{fig:task}
    \vspace{-0.3cm}
\end{figure*}
\begin{figure*}[t]
    \centering
    \begin{subfigure}[b]{0.32\linewidth}
        \includegraphics[width=\linewidth]{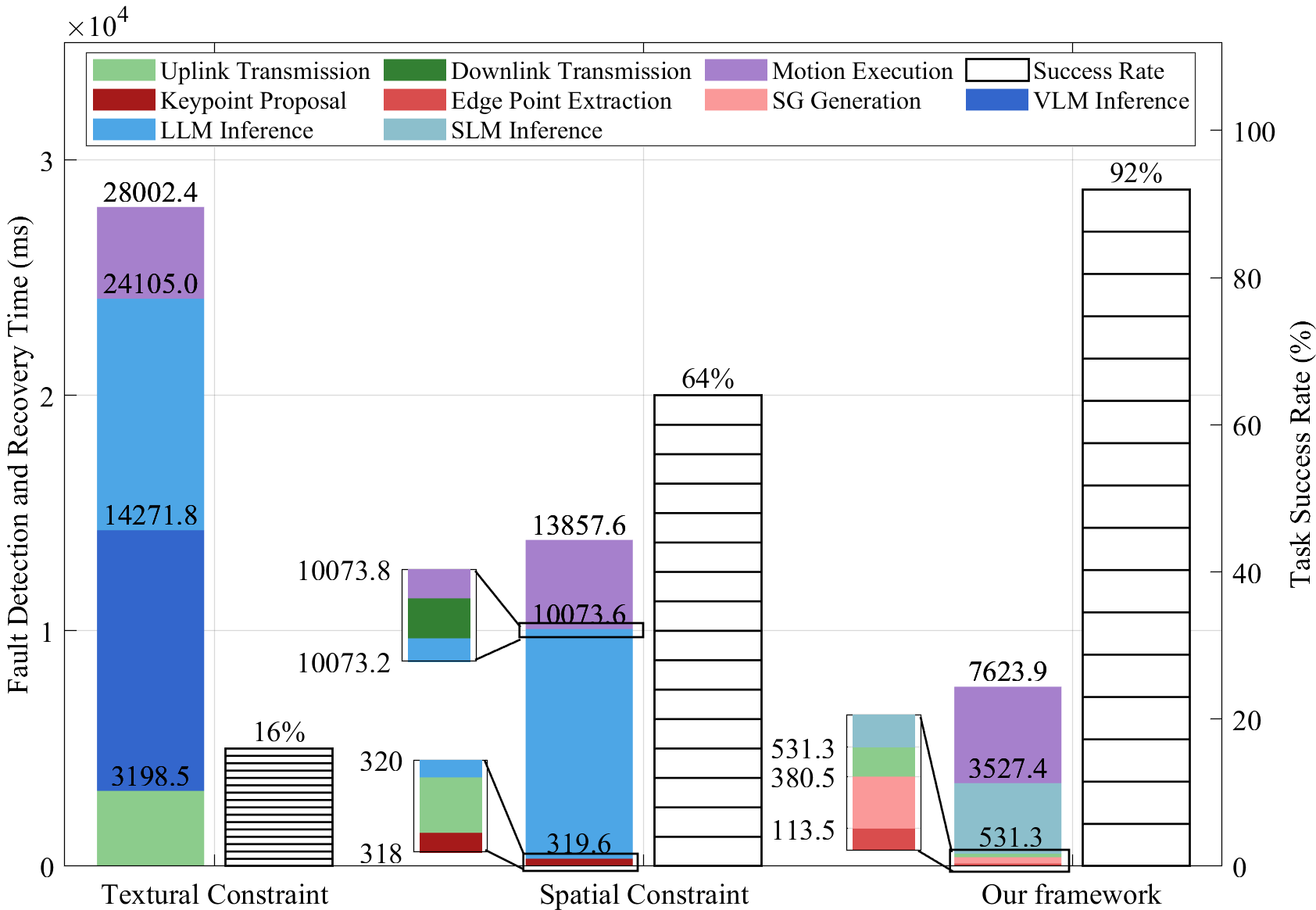}
        \caption{Workpiece Sorting.}
        \label{fig:s_place_load}
    \end{subfigure}
    \begin{subfigure}[b]{0.32\linewidth}
        \includegraphics[width=\linewidth]{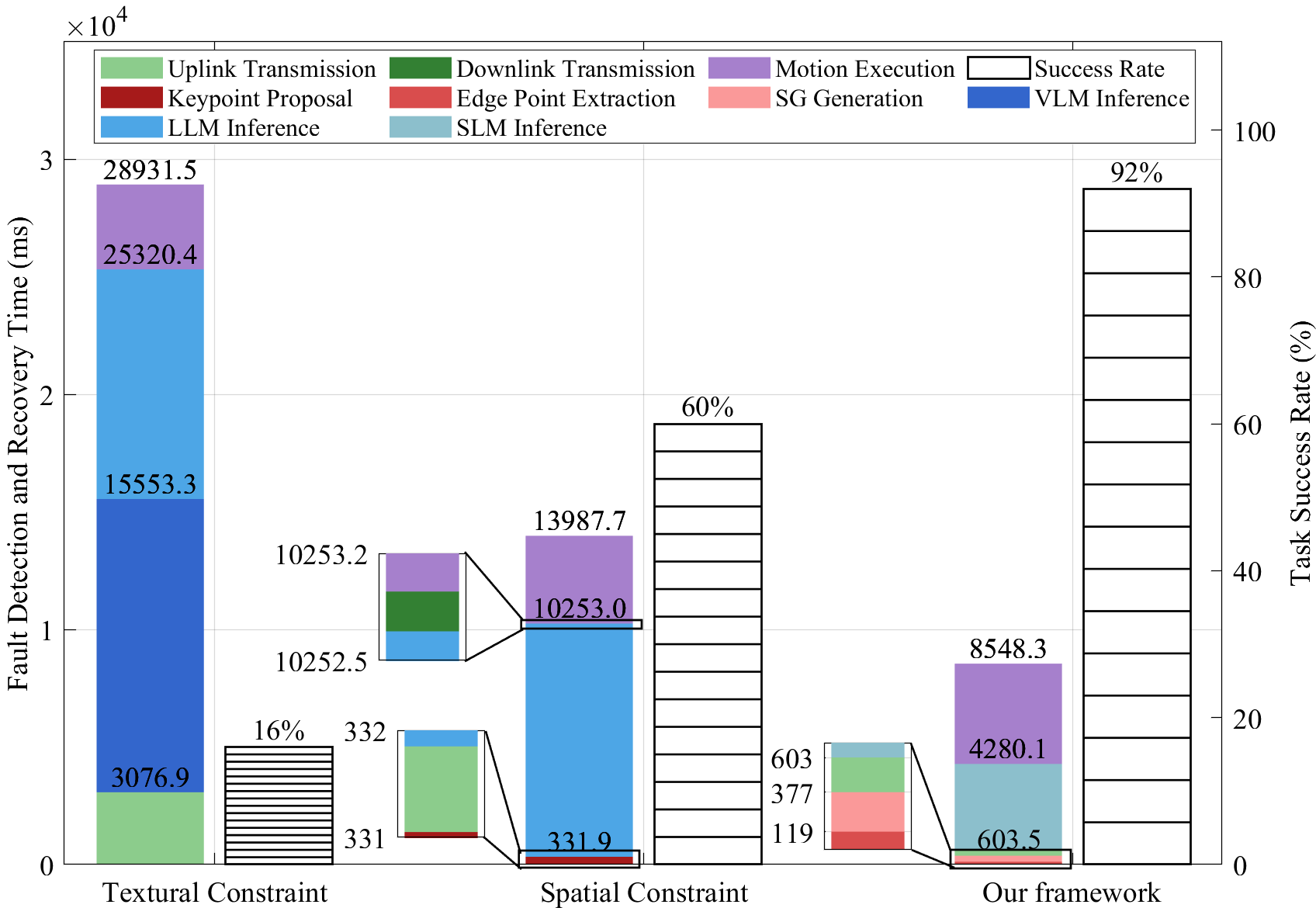}
        \caption{Grocery Packing.}
        \label{fig:s_place_angle}
    \end{subfigure}
    \begin{subfigure}[b]{0.32\linewidth}
        \includegraphics[width=\linewidth]{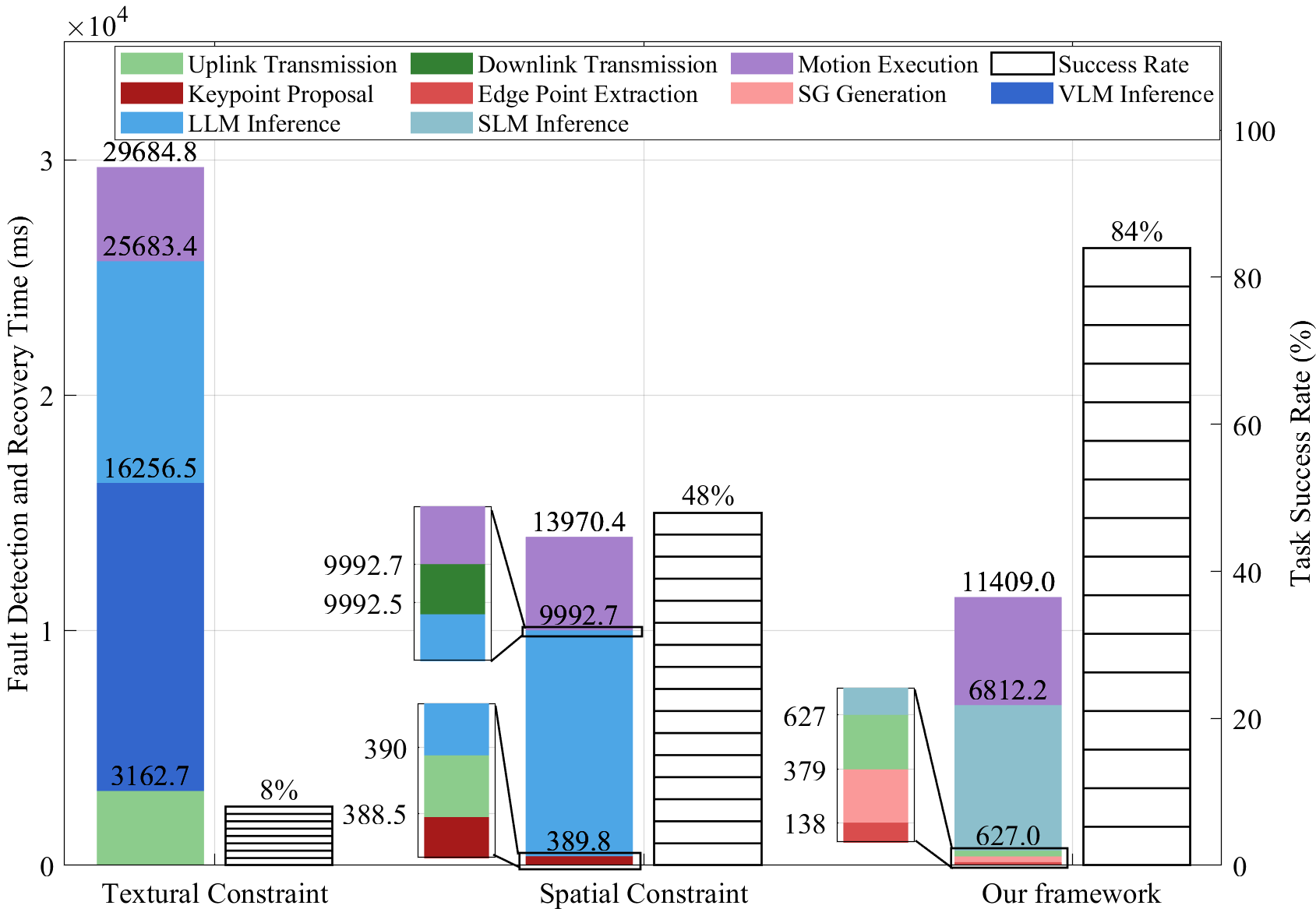}
        \caption{Parcel Palletising.}
        \label{fig:s_place_vel}
    \end{subfigure}
    \caption{\small Comparison of FDR time and task success rate between SOTA and our proposed GoC frameworks in handling motion-level faults.}
    \label{fig:motion}
    \vspace{-0.3cm}
\end{figure*}

\vspace{-0.3cm}
\subsection{Task-level Fault Detection and Recovery}
Fig.~\ref{fig:task} plots the FDR time and task success rate of the SOTA frameworks and our GoC framework across the three tasks when encountering task-level faults, where 3D-SG is transmitted and digital twin reconstruction is not triggered in our framework.
It is observed that, compared with the textual constraint-based framework, our GoC framework significantly reduces the fault detection time by $97.6\%$ without compromising the task success rate. This is because it replaces the periodical image transmission and VLM inference by the lightweight 3D-SG generation and fault-triggered transmission.
It is also observed that, compared with the spatial constraint-based framework, although our GoC framework only achieves a 63.7\% reduction in fault detection time, it consistently achieves at least a $12\%$ higher task success rate across all three tasks. This is because our transmitted 3D-SG provides a more intuitive and interpretable spatial representation, which allows the SLM to directly reason about the cause of failure and determine an appropriate recovery strategy.
In contrast, the keypoint-based spatial observations often lack sufficient geometric detail, which makes it challenging for the replanner (i.e., LLM) to perform accurate global spatial reasoning that is necessary for assessing task completion and identifying failures in the task sequence.
Also, it can be seen that, compared with both SOTA frameworks that use LLMs, our fine-tuned SLM further reduces the fault recovery time (i.e., inference time and motion execution time) by $79.5\%$ with varying levels of improvement in task success rates over the three tasks. This not only demonstrates the comparable reasoning performance of our SLM with significantly lower inference latency, but also highlights the overall soundness of our GoC framework in leveraging the 3D-SG representation for FDR. 
That is, the 3D-SG serves as a compact yet expressive representation that contains sufficient task-relevant information to support both accurate fault detection and effective recovery planning.
Overall, our proposed GoC framework, which incorporates 3D-SG-based fault detection and a fine-tuned SLM, outperforms the SOTA frameworks in both time efficiency and recovery reliability when encountering task-level faults, achieving an up to $82.6\%$ reduction in overall FDR time and up to $36\%$ higher task success rate.

\vspace{-0.3cm}
\subsection{Motion-level Fault Detection and Recovery}
Fig.~\ref{fig:motion} plots the FDR time and task success rate of the SOTA frameworks and our GoC framework across the three tasks when encountering motion-level faults, in which edge points are transmitted and digital twin reconstruction is  triggered in our framework.
{It can be seen that both SOTA frameworks suffer from noticeable degradation in task success rate when handling motion-level faults, while our GoC framework remains consistently effective across all three tasks.
On the one hand, the textual constraint-based framework experiences significant communication and inference delays, which means motion-level faults such as collisions are often detected only after they have already occurred.
On the other hand, the keypoint-based observation captured by the spatial constraint-based framework often lacks the necessary geometric fidelity, which means even when object surfaces are already in contact or on a collision path, the system may erroneously interpret the state as normal.
In contrast, by transmitting object edge points and reconstructing a lightweight digital twin at the edge server, our GoC framework reduces the fault detection time by 97.3\% compared to the textual constraint-based framework, and achieves a similar detection latency to the spatial constraint-based framework. Moreover, it preserves critical geometric information necessary for accurate motion refinement, which leads to a 76\% higher task success rate compared to the textual constraint-based framework, and up to a 36\% improvement over the spatial constraint-based framework.}
Importantly, we also observe that the overall FDR time of our framework is over 50\% longer than that in the task-level case. This is because the recovery process involves digital twin verification, which may trigger multiple rounds of interaction with the SLM before a valid recovery motion is confirmed, causing additional inference delay but ultimately improving the task success rate.
Overall, our proposed GoC framework, which incorporates digital twin reconstruction based on edge points, performs better than the SOTA frameworks when encountering motion-level faults, resulting in a maximum 72.7\% reduction in FDR time and a maximum 76\% higher task success rate.

\begin{figure*}[t]
    \centering
    \begin{subfigure}[b]{0.32\linewidth}
        \includegraphics[width=\linewidth]{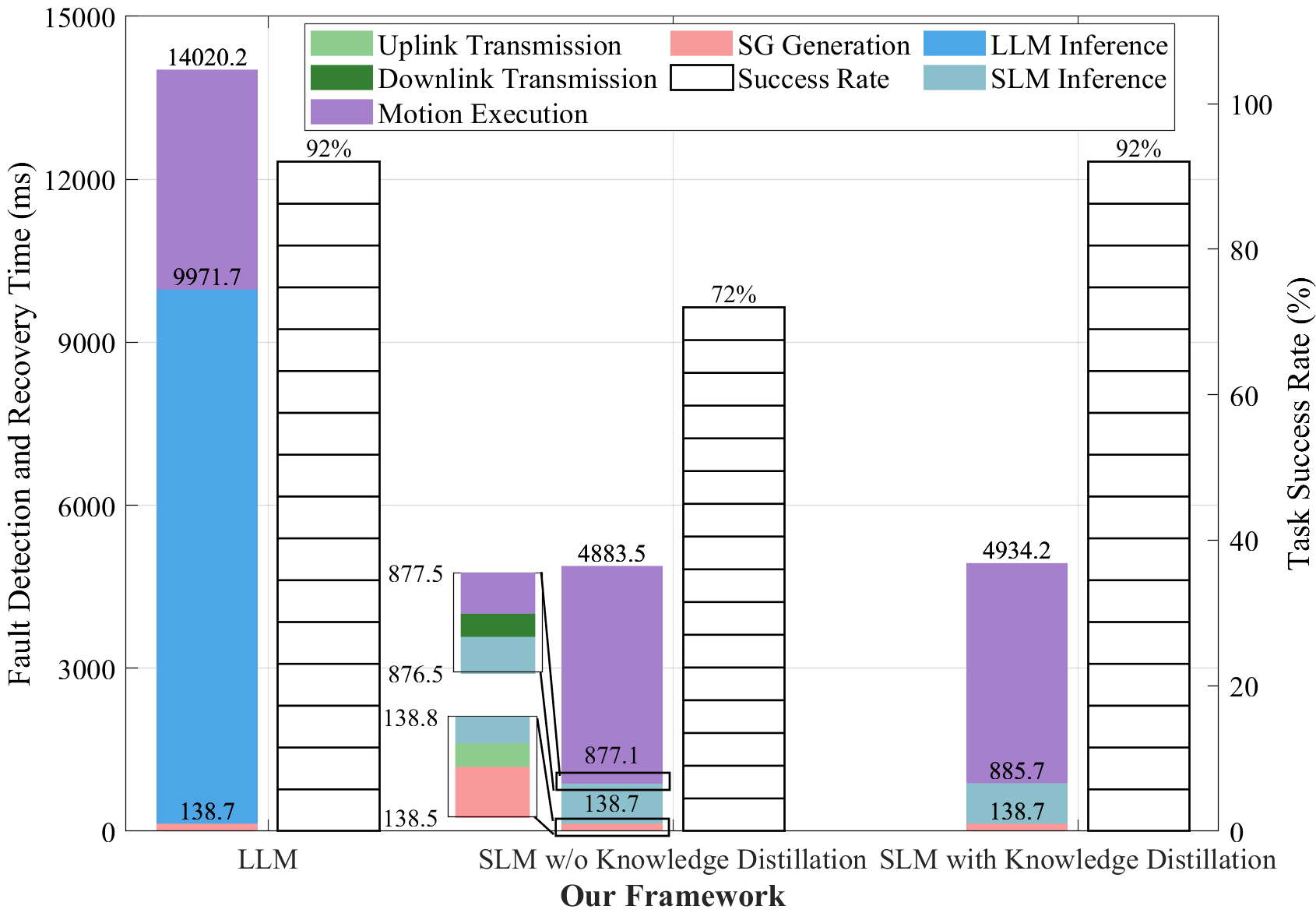}
        \caption{SLM fine-tuning modules.}
        \label{fig:ablation_SLM}
    \end{subfigure}
    \begin{subfigure}[b]{0.32\linewidth}
        \includegraphics[width=\linewidth]{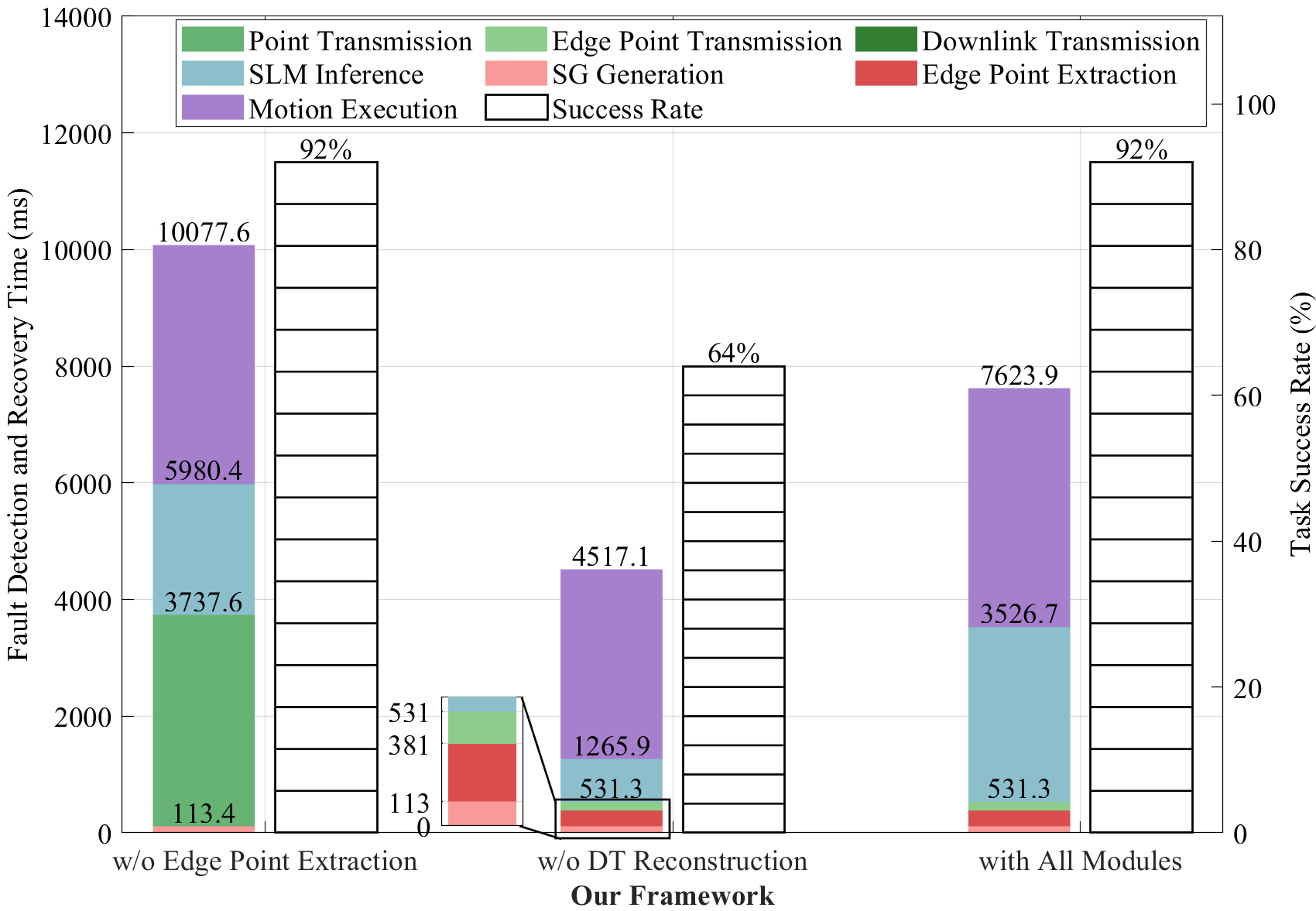}
        \caption{Edge point extraction and digital twin  modules.}
        \label{fig:ablation_dt}
    \end{subfigure}
        \begin{subfigure}[b]{0.32\linewidth}
        \includegraphics[width=\linewidth]{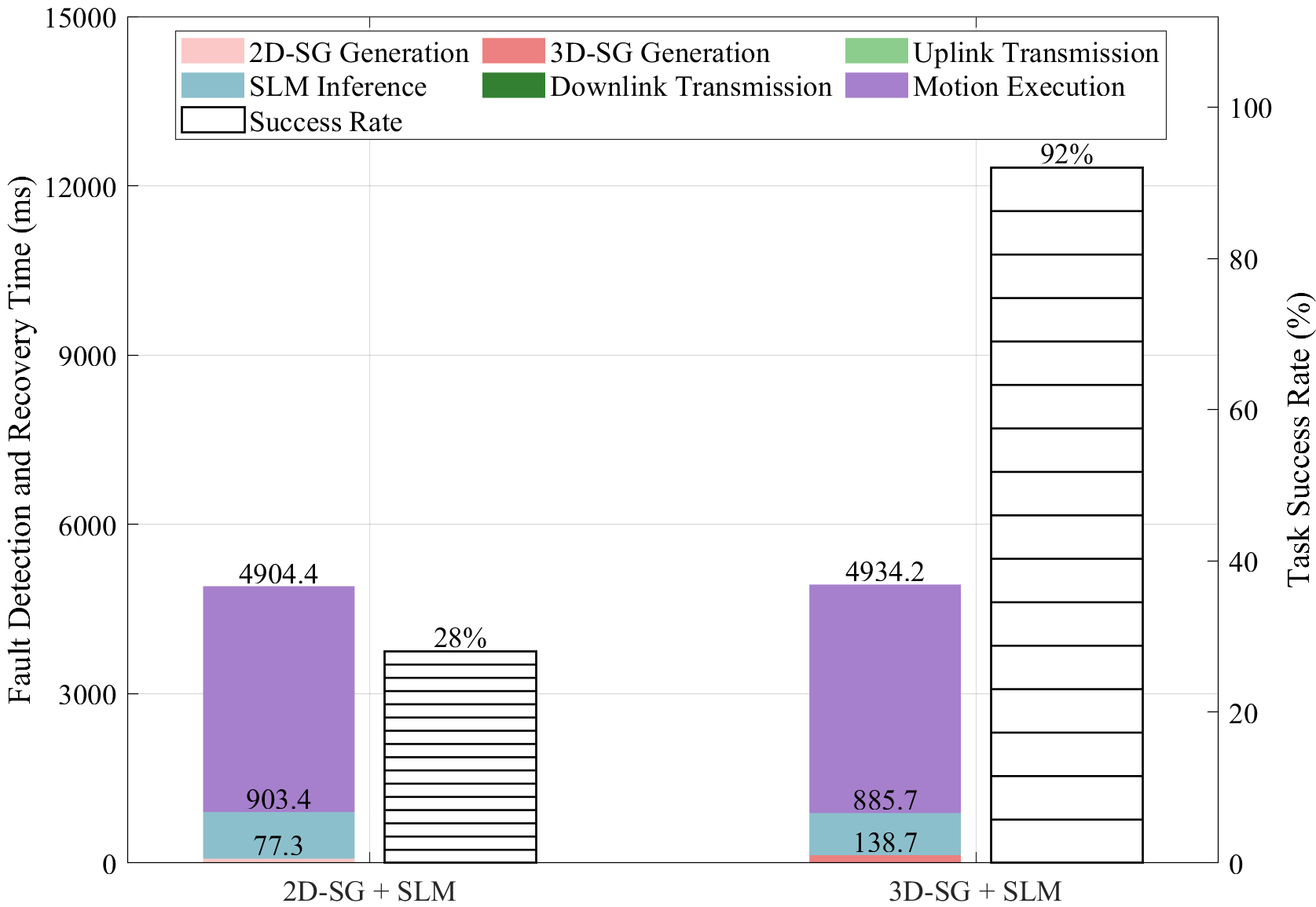}
        \caption{{3D-SG extraction modules.}}
        \label{fig:ablation_sg}
    \end{subfigure}
    \caption{\small Ablation Study of our proposed  GoC framework.}
    \label{fig:ablation}
    \vspace{-0.5cm}
\end{figure*}

\subsection{Ablation Studies}
In this subsection, we conduct ablation studies to examine the effectiveness of our SLM fine-tuning module (shown in Fig.\ref{fig:ablation_SLM}), our edge point extraction and digital twin reconstruction modules (shown in Fig.\ref{fig:ablation_dt}), {as well as our 3D-SG extraction module (shown in Fig.\ref{fig:ablation_sg}).}

Fig.~\ref{fig:ablation_SLM} plots the FDR time and task success rate of our GoC framework when using a LLM, a SLM with fine-tuning but without knowledge distillation, and a SLM with both fine-tuning and knowledge distillation for fault recovery\footnote{We do not include the vanilla SLM without fine-tuning in the comparison, as it consistently failed to generate  recovery actions.}.
It can be seen that, compared with LLM, the use of SLM reduces the fault recovery time by $92\%$, but also results in a drop in inference performance (i.e., success rate) from $92\%$ to $72\%$ due to constrained reasoning capability and limited generalization. However, our task-specific knowledge distillation mitigates this issue and improves the task success rate to similar level of LLM by transferring knowledge from the LLM to the SLM and enhances its generalization in challenging or previously unseen conditions, without increasing the inference delay.
Finally, it can be concluded that our SLM with task-specific fine-tuning and knowledge distillation can ultimately achieve reasoning performance close to that of the LLM, but with a $92\%$ reduction in inference time.

Fig.~\ref{fig:ablation_dt} plots the FDR time and task success rate of our GoC framework under three settings: 
(1) with digital twin reconstruction, (2) with edge point extraction, and (3) with both edge point extraction and digital twin reconstruction.
We first observe that enabling the edge point extraction module significantly reduces the wireless transmission latency by 95.8\% without any degradation in task success rate.
Instead of transmitting the complete scene point cloud, our goal-oriented communication transmits more informative edge points that retain the most critical geometric features necessary for trajectory planning, which again validates the effectiveness of our defined semantic representation tailored for the robotic FDR task.
We also observe that enabling the digital twin reconstruction module improves the task success rate by 28\%, but also increases the SLM inference time by 75\%.
This is because the reconstructed digital twin allows us to simulate and verify the generated recovery motion before execution, where feedback in failure cases triggers the SLM for motion plan refinement, but also results in additional query rounds and thus longer inference latency.
Overall, we conclude that combining our edge point extraction and digital twin reconstruction modules, we reduce the overall FDR time by 24.3\% with a 28\% improvement in task success rate.

Fig.~\ref{fig:ablation_sg} plots the FDR time and task success rate of our GoC framework when replacing the 3D-SG semantic representation with 2D-SG, where we adopt Faster Region-based Convolutional Neural Network (Faster R-CNN)\cite{2DSG} to generate 2D-SG from raw RGB images, and employ in-context learning to enable the SLM to reason over the 2D-SG input. 
It can be seen that, although the 2D-SG achieves a 44\% lower fault detection time and a similar fault recovery time compared to our 3D-SG, its task success rate drops significantly from 92\% to 28\%. This is because the 2D-SG only captures object relationships on the image-plane and cannot preserve the depth-aware spatial information required for robotic FDR. 
For example, in the workpiece sorting task, when the workpiece overlaps with the robot end-effector in the RGB image, the 2D-SG cannot reliably determine whether the robot is still approaching the workpiece or has already grasped it. As a result, the generated scene graph may provide ambiguous or incorrect relational information to the SLM, which further leads to unreliable recovery motion generation.
In contrast, our 3D-SG explicitly models object relations in 3D space, which allows the SLM to reason more accurately about the actual workspace state. Therefore, the superiority of our proposed 3D-SG does not only lie in faster semantic representation generation and wireless transmission, but also in providing the task-relevant information directly required by the downstream FDR task.
\color{black}

\begin{table*}[t]
\centering
\small
\setlength{\tabcolsep}{3pt}
\caption{{End-to-End Fault Detection Latency Breakdown Table of Local Computing and Edge Computing (Unit: ms).}}
\renewcommand{\arraystretch}{1.3}
{
\begin{tabular}{|>{\raggedright\arraybackslash}p{2.9cm}|c|c|c|c|c|c|c|c|}
\hline
\multicolumn{9}{|c|}{\textbf{Bandwidth: 1 MHz}} \\ \hline
\makecell[c]{Scheme} 
& \makecell[c]{Object Detection\\and Segmentation} 
& \makecell[c]{Back\\Projection} 
& \makecell[c]{3D-SG\\Extraction} 
& \makecell[c]{Edge Point\\Extraction} 
& \makecell[c]{3D-SG\\Transmission} 
& \makecell[c]{Edge Point\\Transmission} 
& \makecell[c]{RGB-D Image\\Transmission} 
& \makecell[c]{End-to-End\\Delay} \\ \hline
\makecell[c]{Local Computing for\\Task-level Fault} 
& 30.81 & 4 & 87.93 & N/A & \textless{}1 & N/A & N/A & 122.8 \\ \hline
\makecell[c]{Local Computing for\\Motion-level Fault} 
& 30.81 & 4 & 87.93 & 267.42 & N/A & 247.18 & N/A & 637.3 \\ \hline
\makecell[c]{Edge Computing} 
& 3.9 & \textless{}1 & 31.62 & 60.3 & N/A & N/A & 5218.9 & 5314.7 \\ \hline
\multicolumn{9}{|c|}{\textbf{Bandwidth: 24 MHz}} \\ \hline
\makecell[c]{Local Computing for\\Task-level Fault} 
& 30.81 & 4 & 87.93 & N/A & \textless{}1 & N/A & N/A & 122.5 \\ \hline
\makecell[c]{Local Computing for\\Motion-level Fault} 
& 30.81 & 4 & 87.93 & 267.42 & N/A & 14.65 & N/A & 404.81 \\ \hline
\makecell[c]{Edge Computing} 
& 3.9 & \textless{}1 & 31.62 & 60.3 & N/A & N/A & 308.98 & 404.8 \\ \hline
\multicolumn{9}{|c|}{\textbf{Bandwidth: 431 MHz}} \\ \hline
\makecell[c]{Local Computing for\\Task-level Fault} 
& 30.81 & 4 & 87.93 & N/A & \textless{}1 & N/A & N/A & 122.5 \\ \hline
\makecell[c]{Local Computing for\\Motion-level Fault} 
& 30.81 & 4 & 87.93 & 267.42 & N/A & 1.28 & N/A & 391.44 \\ \hline
\makecell[c]{Edge Computing} 
& 3.9 & \textless{}1 & 31.62 & 60.3 & N/A & N/A & 26.92 & 122.74 \\ \hline
\end{tabular}}
\label{tab:end-to-end}
\end{table*}

\begin{figure}
\centering
\includegraphics[width=0.9\linewidth]{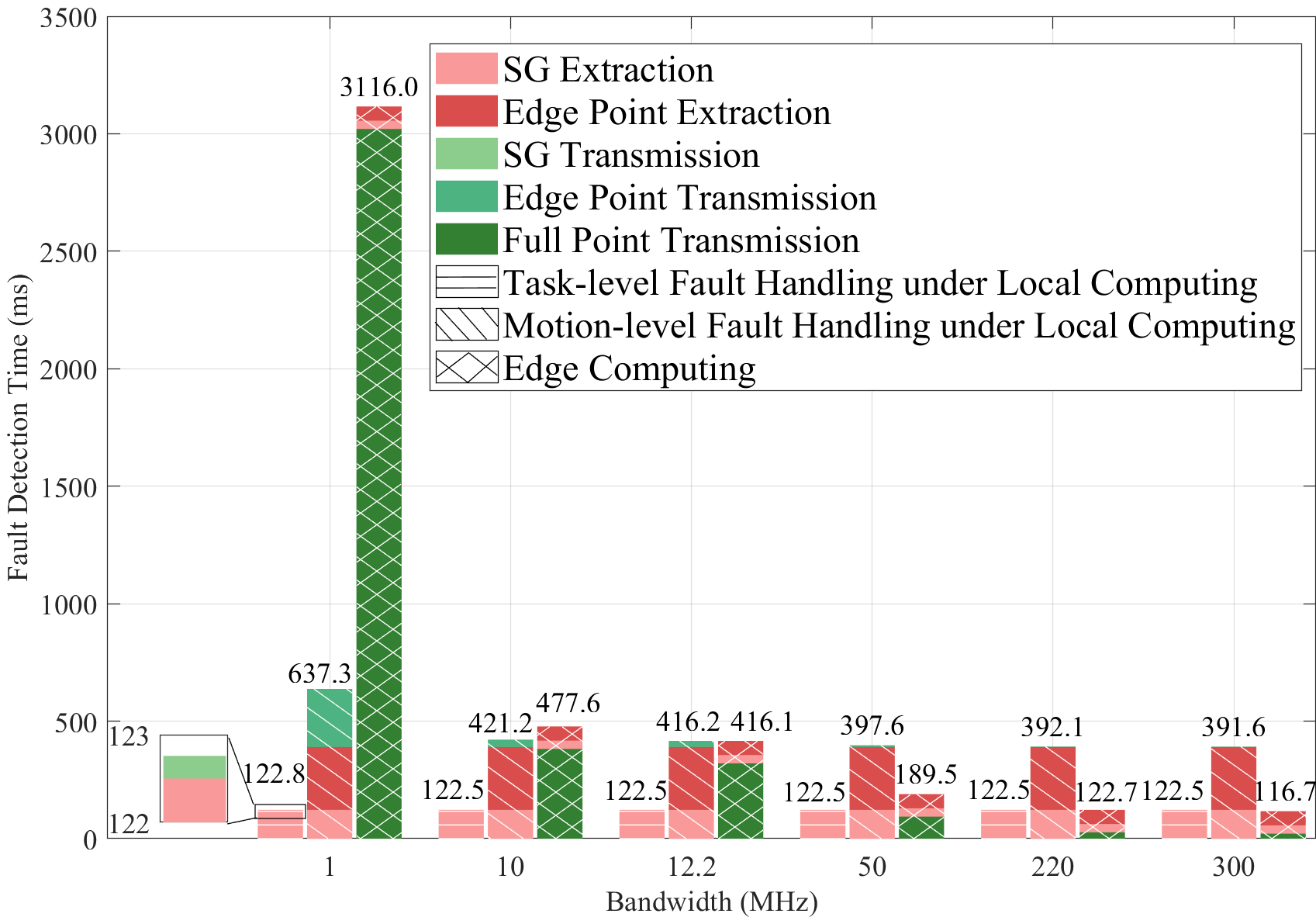}
\caption{Effectiveness of our proposed adaptive offloading scheme.}
\label{fig:offload}
\vspace{-0.3cm}
\end{figure}

\subsection{Adaptive Offloading Scheme}
Fig.~\ref{fig:offload} plots the fault detection time of our proposed GoC framework under different wireless bandwidths, comparing three computation strategies: local computing for task-level faults, local computing for motion-level faults, and edge computing for both types of faults with full point cloud transmission.
It can be seen that when the available bandwidth is low (below 12.2 MHz), both motion-level and task-level faults should be handled via local computing, with only the extracted semantic representations (i.e., 3D-SG or edge points) uploaded to the edge server through the uplink. This is because the limited bandwidth leads to excessive transmission latency for raw point cloud data.
It can also be seen that when the available bandwidth is in the medium range (between 12.2 MHz and 220 MHz), a hybrid offloading strategy becomes preferable, where task-level faults should still be handled via local computing, while motion-level faults can be processed at the edge server. This is because the semantic representation extraction time remains constant regardless of bandwidth, whereas the reduction in transmitting edge points makes edge processing feasible.
It can also be observed that when the available bandwidth is sufficiently high (beyond 220 MHz), offloading both task-level and motion-level faults to the edge server is an optimal solution.
In this case, transmitting complete point clouds can fully exploit edge computing resources for faster fault detection.

{Tab. \ref{tab:end-to-end} reports the end-to-end fault detection latency breakdown of our proposed GoC framework under different wireless bandwidths and the three offloading schemes, but the raw RGB-D images with uint16 depth were used to replace the full point cloud in edge computing scheme. 
The results exhibit a trend consistent with Fig. \ref{fig:offload}: 1) when the available bandwidth is below 23.7 MHz, local computing is preferred; 2) when the bandwidth is between 23.7 MHz and 430.8 MHz, a hybrid offloading strategy is preferable, where task-level faults are handled by local computing while motion-level faults are processed at the edge server; 3) when the bandwidth exceeds 430.8 MHz, edge computing becomes the optimal choice. It is also observed that the bandwidth thresholds allowed to transmit RGB-D  images are much higher than those in Fig. \ref{fig:offload}, because the data size of raw RGB-D images is significantly larger than that of the full point cloud. This indicates that, as the size of raw transmitted data increases, the benefit of our adaptive offloading scheme becomes more pronounced.}

\section{Conclusion} \label{sec:conclu}
In this work, we proposed a Goal-oriented  Communication (GoC) framework for fast and robust robotic fault detection and recovery (FDR), to minimise the FDR time while maximising the robotic task success rate.
Unlike the state-of-the-art (SOTA) frameworks that design communication, computation, and control modules independently of the specific objectives of FDR, our GoC framework jointly designed three modules to serve the downstream FDR task.
For communication, we innovatively defined the 3D scene graph (3D-SG) as the semantic representation tailored for fault detection. The 3D-SG is extracted from noisy and partial point clouds using our designed representation extractor, and is transmitted only upon fault detection, instead of periodically transmitting raw visual data as in SOTA frameworks.
For computing, we replaced the large language models (LLMs) used in SOTA frameworks with an expert small language model (SLM) tailored to FDR. The SLM is fine-tuned via Low-Rank Adaptation (LoRA) for low-latency inference and further enhanced through knowledge distillation to improve reasoning accuracy.
For control, we proposed a lightweight task-oriented digital twin reconstruction module that refines the robotic trajectories generated by SLM when  high-precision control is required, where only contours of task-relevant object are reconstructed.
Simulations shown that our GoC framework reduces the average FDR time by up to 82.6\% compared to SOTA frameworks, and achieves up to a 76\% higher task success rate.


\bibliographystyle{IEEEtran}
\bibliography{IEEEabrv,FDR}

\end{document}